\pdfoutput=1

\documentclass[11pt]{article}

\usepackage{microtype}
\usepackage{graphicx}
\usepackage{subfigure}
\usepackage{booktabs} 

\usepackage{amsmath} 
\usepackage{multirow}

\usepackage{hyperref}
\usepackage{url}
\usepackage[utf8]{inputenc} 
\usepackage[T1]{fontenc}    
\usepackage{hyperref}       
\usepackage{url}            
\usepackage{booktabs}       
\usepackage{amsfonts}       
\usepackage{nicefrac}       
\usepackage{microtype}      
\usepackage{makecell}
\usepackage{graphicx}
\usepackage{amsmath}
\usepackage{multirow}
\usepackage{wrapfig}
\usepackage{algorithm}
\usepackage{algpseudocode}
\usepackage{wrapfig}
\usepackage{float}
\usepackage{graphicx}
\usepackage{caption}
\usepackage{color}
\usepackage{colortbl}
\usepackage{threeparttable}

\newcommand{\EQ}{\begin{eqnarray}}

\newcommand{\EN}{\end{eqnarray}}
\newcommand{\EQQ}{\begin{eqnarray*}}
\newcommand{\ENN}{\end{eqnarray*}}
\newcommand{\overlineray}{\begin{array} }

\newcommand{\barray}{\begin{array} }
\newcommand{\earray}{\end{array}}

\newcommand{\bremark}{\begin{remark} }
\newcommand{\eremark}{\end{remark}}
\newcommand{\btheorem}{\begin{theorem}}
\newcommand{\etheorem}{\end{theorem}}
\newcommand{\blemma}{\begin{lemma}}
\newcommand{\elemma}{\end{lemma}}
\newcommand{\bassumption}{\begin{assumption} }
\newcommand{\eassumption}{\end{assumption}}
\newcommand{\bcorollary}{\begin{corollary} }
\newcommand{\ecorollary}{\end{corollary}}
\newcommand{\bdefinition}{\begin{definition} }
\newcommand{\edefinition}{\end{definition}}
\newcommand{\bproposition}{\begin{proposition}}
\newcommand{\eproposition}{\end{proposition}}

\newcommand{\balgorithm}{\medskip\begin{algorithm} \rm}
\newcommand{\ealgorithm}{ \hfill \rule{1mm}{2mm}\medskip
\end{algorithm} }

\usepackage{microtype}
\usepackage{graphicx}
\usepackage{subfigure}
\usepackage{booktabs} 
\usepackage{paralist}
\usepackage{bbding}

\usepackage[preprint]{acl}

\usepackage{times}
\usepackage{latexsym}

\usepackage[T1]{fontenc}

\usepackage[utf8]{inputenc}

\usepackage{microtype}

\usepackage{inconsolata}

\usepackage{graphicx}

\title{Linear-MoE: Linear Sequence Modeling Meets Mixture-of-Experts}

\author{
 \textbf{Weigao Sun$^{1}$}\textsuperscript{\dag},
 \textbf{Disen Lan$^{1,2}$},
 \textbf{Tong Zhu$^{3}$},
 \textbf{Xiaoye Qu$^{1}$},
 \textbf{Yu Cheng$^{4}$}
\\
 \textsuperscript{1}Shanghai AI Laboratory,
 \textsuperscript{2}South China University of Technology,
 \textsuperscript{3}Soochow University,
\\
 \textsuperscript{4}The Chinese University of Hong Kong
\\
}
\newcommand\blfootnote[1]{%
\begingroup
\renewcommand\thefootnote{}\footnote{#1}%
\addtocounter{footnote}{-1}%
\endgroup
}

\begin{document}
\maketitle

\blfootnote{\dag\ Project lead. Corresponding to Weigao Sun (sunweigao@outlook.com). Work done during Disen Lan’s internship at Shanghai AI Laboratory.}

\begin{abstract}

Linear Sequence Modeling (LSM) like linear attention, state space models and linear RNNs, and Mixture-of-Experts (MoE) have recently emerged as significant architectural improvements. In this paper, we introduce Linear-MoE, a production-level system for modeling and training large-scale models that integrate LSM with MoE. Linear-MoE leverages the advantages of both LSM modules for linear-complexity sequence modeling and MoE layers for sparsely activation, aiming to offer high performance with efficient training. The Linear-MoE system comprises: 1) Modeling subsystem, which provides a unified framework supporting all instances of LSM. and 2) Training subsystem, which facilitates efficient training by incorporating various advanced parallelism technologies, particularly Sequence Parallelism designed for Linear-MoE models. Additionally, we explore hybrid models that combine Linear-MoE layers with standard Transformer-MoE layers with its Sequence Parallelism to further enhance model flexibility and performance. Evaluations on two model series, A0.3B-2B and A1B-7B, demonstrate Linear-MoE achieves efficiency gains while maintaining competitive performance on various benchmarks, showcasing its potential as a next-generation foundational model architecture. Code: \url{https://github.com/OpenSparseLLMs/Linear-MoE}.
\end{abstract}

\section{Introduction}


Mixture-of-Experts (MoE) \cite{Jacobs_Jordan_Nowlan_Hinton_1991,qu2024llama} architectures have gained widespread adoption in cutting-edge models within industry, with prominent examples including Gemini-1.5~\citep{reid2024gemini} and the reported use of MoE in GPT-4~\citep{gpt4}. Other notable large models incorporating MoE techniques include Mixtral~\citep{jiang2024mixtral}, DeepSeek V2~\citep{liu2024deepseek}, Qwen2~\citep{qwen2}, JetMoE~\citep{shen2024jetmoe}, Jamba~\citep{team2024jamba}, and OLMoE~\citep{muennighoff2024olmoe}.

Most advances on MoE studies primarily concentrate on modifying the routing mechanism or expert layers, while typically keeping the attention layers unchanged \cite{zhu2024llama}. These attention layers commonly rely on the softmax self-attention mechanism introduced in the Transformer architecture~\citep{vaswani2017attention}. The softmax-based self-attention has proven to be highly effective for sequence modeling tasks across various data types.
However, a significant limitation of this mechanism is its computational complexity, which grows quadratically with the input sequence length. This complexity can lead to substantial computational costs, especially during training, making it a challenge for models need to handle long sequences efficiently.


Linear sequence modeling (LSM) has recently gained significant attention due to its impressive efficiency in both training and inference. These methods function similarly to recurrent neural networks (RNNs) with matrix-valued hidden states, allowing them to achieve linear-time training and constant-memory inference. This efficiency is largely due to the fact that LSM techniques bypass the computation of attention scores and eliminate the need for maintaining a key-value (KV) cache. 
There are three primary approaches to linear sequence modeling: linear attention~\citep{katharopoulos2020transformers}, state space models  (SSM)~\citep{gu2023mamba,dao2024transformers,hu2024timessm,waleffe2024empirical}, and linear RNN~\citep{peng-etal-2023-rwkv,peng2024eagle,qin2024hgrn2}. Linear attention is a variation of the traditional softmax attention mechanism, replacing the exponential kernel with a simpler dot product between key and query vectors, which enables the use of the right-product kernel trick to reduce computational complexity. SSM approaches, such as Mamba and Mamba2, stem from control theory and represent sequence modeling as dynamic systems. Meanwhile, linear RNN methods address the limitations of traditional RNNs in modeling long contexts by enabling parallel training of RNN models. 
These different methods, linear attention, SSM, and linear RNN, share a common mathematical foundation and exhibit similar performance on sequence modeling  tasks~\cite{dao2024transformers,peng2024eagle,qin2024unlocking,yang2024parallelizing}. In fact, they all employ a unified recurrence framework expressed as $\mathbf{M}_s = \mathbf{M}_{s-1} + \widehat{\mathbf{M}}_s$, where $\mathbf{M}_s$ denotes the memory state and $\widehat{\mathbf{M}}_s$ represents the incremental memory update at the $s$-th token.


In this paper, we introduce Linear-MoE, a production-level system designed for modeling and training of large-scale MoE models with LSM modules integrated. The Linear-MoE system is composed of two key subsystems: Modeling and Training. The Modeling subsystem provides a unified linear sequence modeling framework for Linear-MoE models. It supports three main types of LSM methods: linear attention, state space model (SSM), and linear RNN. For each type, multiple instances are implemented under a unified formulation. While the Training subsystem is designed to achieve efficient training of Linear-MoE models on modern accelerators. In addition to supporting state-of-the-art training techniques, we incorporate a specialized Sequence Parallelism (SP) technique for LSM modules, which is particularly effective for handling extremely long input sequences on Linear-MoE architecture. Importantly, the system is designed to be extensible, enables more advanced sequence modeling methods or training techniques integrated in the future. Furthermore, we also explore efficient modeling and training for hybrid Linear-MoE models, which combine Linear-MoE layers with standard Transformer-MoE layers. For hybrid models, we introduce an SP method that employs distinct computational and communication strategies tailored to the different types of layers. 



Our contributions can be summarized as follows: 

\begin{compactitem}

\item {\textit{Production-level System.}} We introduce Linear-MoE, a production-level system designing for efficient modeling and training of large-scale MoE models with LSM modules integrated.

\item {\textit{Modeling \& Training Subsystems.}} The Linear-MoE system is composed of two subsystems: \textit{Modeling} and \textit{Training}. We provide unified linear sequence modeling formulation to support various LSM modules with MoE layers, as well as state-of-the-art training techniques for efficient large-scale model training, especially on long-context inputs.

\item {\textit{Experimental Validation.}} In empirical studies, we pretrain two series of Linear-MoE models from scratch on the public SlimPajama corpus. Extensive experiments validate the training and inference efficiency of our system framework, as well as the performance of Linear-MoE architecture on various downstream tasks.

\end{compactitem}


\begin{table*}[t]
    \centering
    \small    
    \caption{\textbf{Instances of Linear Sequence Modeling Methods.} All instances listed follow the unified formulation in Eq.~(\ref{eq:general_form}). Here, $a \in \mathbb{R}$, $a_s \in \mathbb{R}$, $\mathbf{a}_s \in \mathbb{R}^{d}$, $\mathbf{A} \in \mathbb{R}^{d \times d}$, $\mathbf{A}_s \in \mathbb{R}^{d \times d}$ represents a fixed constant, a time-dependent scalar, a time-dependent vector, a time-independent matrix, and a time-dependent matrix, respectively. Note that the same notation may denote different variables in different instances.}
    \setlength{\tabcolsep}{3mm}
    \begin{threeparttable}
        \begin{tabular}{llll}
        \toprule
            \textbf{LSM Method} & \textbf{Instance}  & \textbf{Recurrent Update Rule} & \textbf{Parameter}  \\  \midrule
            Linear Attention & BLA & $\mathbf{M}_s = \mathbf{M}_{s-1} + \mathbf{k}_s^\top \mathbf{v}_s$ & $\backslash$  \\
             & Lightning Attn & $\mathbf{M}_s = a \mathbf{M}_{s-1} + \mathbf{k}_s^\top \mathbf{v}_s$ & $a \in \mathbb{R}$ \\
             & RetNet & $\mathbf{M}_s = a \mathbf{M}_{s-1} + \mathbf{k}_s^\top \mathbf{v}_s$ & $a \in \mathbb{R}$ \\
             & GLA & $\mathbf{M}_s = \text{diag}\{\mathbf{a}_s\} \mathbf{M}_{s-1} + \mathbf{k}_s^\top \mathbf{v}_s$ & $\mathbf{a}_s \in \mathbb{R}^{d}$ \\
             & DeltaNet & $\mathbf{M}_s = (\mathbf{I}-a_s \mathbf{k}_s^\top \mathbf{k}_s) \mathbf{M}_{s-1} + b_s \mathbf{k}_s^\top \mathbf{v}_s$ & $a_s, b_s \in \mathbb{R}$ \\
             & Rebased & $\mathbf{M}_s = \mathbf{M}_{s-1} + \phi(\mathbf{k}_s)^\top \mathbf{v}_s$ & $\backslash$ \\
             & GFW & $\mathbf{M}_s = \mathbf{A}_s \odot \mathbf{M}_{s-1} + \mathbf{k}_s^\top \mathbf{v}_s$ & $\mathbf{A}_s \in \mathbb{R}^{d\times d}$ \\
             & GateLoop & $\mathbf{M}_s = \mathbf{A}_s \odot \mathbf{M}_{s-1} + \mathbf{k}_s^\top \mathbf{v}_s$ & $\mathbf{A}_s \in \mathbb{R}^{d\times d}$ \\
             & Gated DeltaNet & $\mathbf M_s = a_s(\mathbf I - \mathbf k_s^\top \mathbf k_s) \mathbf M_{s-1} + b_s \mathbf k_s^\top \mathbf v_s$ & $a_s, b_s \in \mathbb{R}$ \\
             & TTT & $\mathbf M_s = \mathbf M_{s-1} + b_s \nabla l(\mathbf M_{s-1};\mathbf k_s, \mathbf v_s) $ & $b_s \in \mathbb{R}$ \\
             & Titans & $\mathbf M_s = a_s \mathbf M_{s-1} + b_s \nabla l(\mathbf M_{s-1};\mathbf k_s, \mathbf v_s) $ & $a_s, b_s \in \mathbb{R}$ \\
             \midrule
            SSM \tnote{*}
            & S4 & $\mathbf{M}_s = \exp(-(\mathbf{a} \mathbf{1}^\top)\mathbf{A}) \odot \mathbf{M}_{s-1} + (\mathbf{a} \mathbf{1}^\top)\mathbf b^\top \mathbf{v}_s$ & $\mathbf{a, b} \in \mathbb{R}^d, \mathbf{A} \in \mathbb{R}^{d \times d}$ \\
            & Mamba & $\mathbf{M}_s = \exp(-(\mathbf{a}_s \mathbf{1}^\top)\mathbf{A}_s) \odot \mathbf{M}_{s-1} + (\mathbf{a}_s \mathbf{1}^\top)\mathbf k_s^\top \mathbf v_s$ & $\mathbf{a}_s \in \mathbb{R}^d, \mathbf{A}_s \in \mathbb{R}^{d \times d}$ \\
            & Mamba2 & $\mathbf{M}_s = \exp(-{a}{b}_s) \odot \mathbf{M}_{s-1} + b_s \mathbf{k}_s^\top \mathbf{v}_s$ & $a, b_s \in \mathbb{R}$ \\
            \midrule
             & HGRN2 & $\mathbf{M}_s = \text{diag}\{\mathbf{a}_s\} \mathbf{M}_{s-1} + (1-\mathbf{a}_s)^\top \mathbf{v}_s$ & $\mathbf{a}_s \in \mathbb{R}^{d}$ \\
            Linear RNN & RWKV6 & $\mathbf{M}_s = \text{diag}\{\mathbf{a}_s\} \mathbf{M}_{s-1} + \mathbf{k}^\top_s \mathbf{v}_s$ & $\mathbf{a}_s \in \mathbb{R}^{d}$ \\
             & RWKV7 & $\mathbf{M}_s = \text{diag}\{\mathbf{a}_s\} \mathbf{M}_{s-1} + \nabla l (\mathbf{M}_{s-1}; \mathbf k_s, \mathbf v_s)$ & $\mathbf{a}_s \in \mathbb{R}^{d}$ \\
        \bottomrule             
        \end{tabular}
 \begin{tablenotes}
\item [*] \begin{small}
    For both S4 and Mamba, the Euler Discretization \citep{gu2020hippo} is applied, such that $\mathbf{\bar{B}=\Delta B}$, and the unprojected $\mathbf x_s$ is denoted as $\mathbf v_s$ for consistency with other formulas.
\end{small}
\end{tablenotes}
\end{threeparttable}
\label{tab:instances}
\end{table*}

\section{Linear-MoE System}


\subsection{Modeling}

\subsubsection{Unified Linear Sequence Modeling}



The standard softmax attention~\citep{vaswani2017attention}, commonly used in transformer models, whose parallel computation form during training can typically be expressed as:
\begin{equation}
    \mathbf{O} = \operatorname{Softmax}(\mathbf{Q} \mathbf{K}^\top) \mathbf{V}.
\end{equation}
Here, the matrices $\mathbf{Q}, \mathbf{K}, \mathbf{V}, \mathbf{O} \in \mathbb{R}^{N\times d}$ correspond to the query, key, value, and output matrices, respectively. The matrices $\mathbf{Q}, \mathbf{K},$ and $\mathbf{V}$ are linear projections of the input matrix $\mathbf{X} \in \mathbb{R}^{N\times d}$, defined as $\mathbf{Q} = \mathbf{X} \mathbf{W}_Q$, $\mathbf{K} = \mathbf{X} \mathbf{W}_K$, and $\mathbf{V} = \mathbf{X} \mathbf{W}_V$, where $\mathbf{W}_Q, \mathbf{W}_K, \mathbf{W}_V \in \mathbb{R}^{d\times d}$ are learnable weight matrices. Here, $N$ and $d$ represent the sequence length and hidden dimension.


Linear Attention~\citep{katharopoulos2020transformers} as one of the representative LSM methods, has emerged as a viable alternative to traditional softmax attention by implementing two primary modifications. First, it eliminates the $\text{Softmax}(\cdot)$ operation, instead embedding it within a kernel feature map. Second, it leverages the associative property of matrix multiplication, reconfiguring $(\mathbf Q \mathbf K^\top)\mathbf V$ into $\mathbf Q (\mathbf K^\top \mathbf V)$. 
These changes reduce both the computational and memory complexity from $\mathbb{O}(N^2 d)$ to $\mathbb{O}(N d^2)$. This approach is frequently referred to as the right-product kernel trick, as it prioritizes matrix product on the right side.

While during inference, both softmax self-attention and linear attention handle a single token at each iteration. Given the $s$-th token $\mathbf x_s \in \mathbb{R}^{1\times d}$, softmax self-attention computes requiring the storage of an expanding set of keys $\{k_1, \cdots, k_s\}$ and values $\{v_1, \cdots, v_s\}$ i.e., the KV cache, which leads to a significant memory burden when dealing with long input sequences:

\begin{equation}
\begin{aligned}
&\mathbf q_s, \mathbf k_s, \mathbf v_s = \mathbf x_s \mathbf W_Q, \mathbf x_s \mathbf W_K, \mathbf x_s \mathbf W_V, \\
&\mathbf o_s = \frac{\sum_{i=1}^{s} \exp(\mathbf q_s {\mathbf k_i}^\top)\mathbf v_i}{\sum_{i=1}^{s} \exp(\mathbf q_s \mathbf k_i^\top)}.
\label{eq: attention inference}
\end{aligned}
\end{equation}
Linear attention replaces the term $\exp(\mathbf{q}_s \mathbf{k}_i^{\top})$ with a kernel $k(\mathbf{x}, \mathbf{y})$ with an associated feature map $\phi$, i.e., $k(\mathbf{x}, \mathbf{y}) = \langle \phi(\mathbf{x}), \phi(\mathbf{y}) \rangle$. This simplifies the calculation of $\mathbf{o}_s$ as
\begin{equation}
    \begin{aligned}
        \mathbf o_s = \frac{\sum_{i=1}^{s} \phi(\mathbf q_s) \phi({\mathbf k_i})^\top \mathbf v_i}{\sum_{i=1}^{s} \phi(\mathbf q_s) \phi(\mathbf k_i)^\top}.
    \end{aligned}
    \label{eq: linear attn inference}
\end{equation}
Letting $\mathbf{M}_s = \sum_{i=1}^s \phi(\mathbf{k}_i)^\top \mathbf{v}_i$ and $\mathbf{z}_s = \sum_{i=1}^{s}\phi(\mathbf{k}_i)^\top$ where $\mathbf{M}_s \in \mathbb{R}^{d\times d}$, $\mathbf{z}_s \in \mathbb{R}^{d\times 1}$, we can rewrite Eq.~(\ref{eq: linear attn inference}) as an RNN:
\begin{equation}
    \begin{aligned}
        \mathbf{M}_s &= \mathbf{M}_{s-1} + \phi(\mathbf{k}_s)^\top \mathbf{v}_s,\\
        \mathbf{z}_s &= \mathbf{z}_{s-1} + \phi(\mathbf{k}_s)^\top,\\
        \mathbf{o}_s &= \frac{\phi(\mathbf{q}_s)\mathbf{M}_s}{\phi(\mathbf{q}_s)\mathbf{z}_s}.
    \end{aligned}
\end{equation}

Follow-up studies on SSM (e.g., Mamba2~\citep{dao2024transformers}) and linear RNNs (e.g., HGRN~\citep{qin2024hgrn2} and RWKV~\citep{peng2024eagle,peng2025rwkv} series), have demonstrated their similarity with linear attention. In fact, recent studies~\citep{qin2024unlocking,yang2024parallelizing} have suggested that linear attention, state space, and linear RNN sequence modeling methods can be expressed within a unified recurrence framework as:
\begin{equation}
\begin{aligned}
\widehat{\mathbf{M}}_{s} &=  f(\mathbf{k}_s^{\top}, \mathbf{v}_s), \\
    \mathbf{M}_{s} &=  \mathbf{\Theta}_s \diamond \mathbf{M}_{s-1} +\widehat{\mathbf{M}}_s.
    \label{eq:general_form}
\end{aligned}
\end{equation}
In this formulation, $\widehat{\mathbf{M}}_s \in \mathbb R^{d\times d}$ represents the memory state corresponding to the $s$-th input, which is a function of $\mathbf{k}_s^{\top}$ and $\mathbf{v}_s$. And $\mathbf{\Theta}_s$ denotes a coefficient matrix that may be time-varying or constant (and also can be a vector or scalar). The operator "$\diamond$" can denote either standard matrix multiplication or Hadamard product. We collect recent LSM method instances which follow the unified formulation in Eq.~(\ref{eq:general_form}) and list them in Table~\ref{tab:instances}.

\begin{figure*}[t]
    \centering
    \vspace{-10mm}
    \hspace{6mm}
    \includegraphics[width=0.85\textwidth]{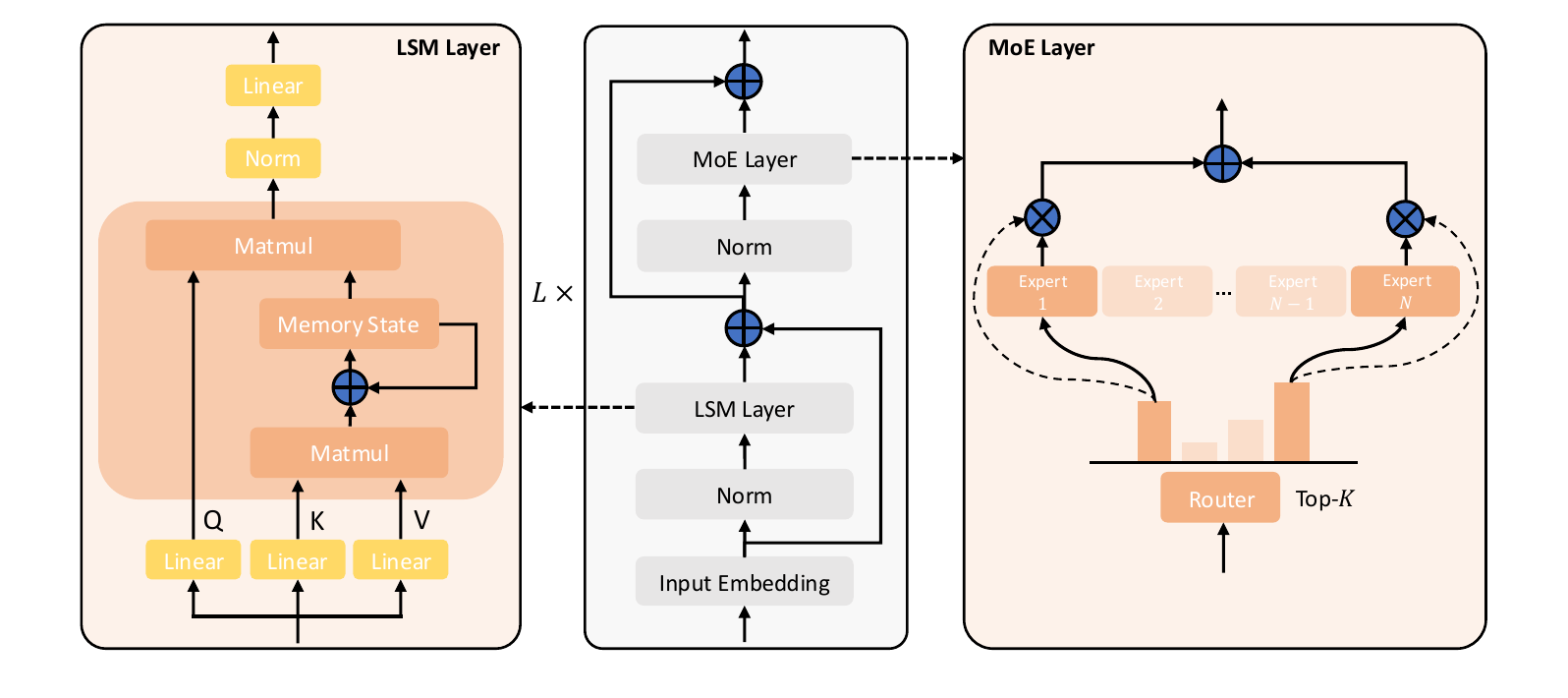}
    \caption{\textbf{Linear-MoE Architecture.} In each Linear-MoE block, there is both an LSM layer and an MoE layer, with each layer preceded by its own normalization layer. The LSM layer is designed as a flexible abstraction of LSM methods, including: linear attention, SSM, and linear RNN, which follows a unified recurrence framework.}
    \vspace{-2mm}
    \label{fig: arch}
\end{figure*}

\subsubsection{Linear-MoE Architecture}

The Linear-MoE architecture is relatively straightforward, consisting of $L \times$ stacked Linear-MoE blocks, as depicted in Fig.~\ref{fig: arch}. Each Linear-MoE block includes an LSM layer and an MoE layer, with a normalization layer preceding each. The LSM layer serves as a generalized structure that supports various LSM methods, specifically, linear attention, SSM, and linear RNN, each encompassing multiple instance methods. Table~\ref{tab:instances} provides an overview of these LSM method instances, unified under a common recurrence framework. This framework highlights key distinctions between various LSM instances, primarily in their handling of the prior-step memory state $\mathbf{M}_{s-1}$ and the computation of the incremental memory state $\widehat{\mathbf{M}}_s$. For the MoE layers, we retain the standard mechanisms of sparse expert activation and routing, as employed in SOTA open-source MoE models. These mechanisms are essential for maintaining an optimal balance between model performance and computational efficiency.

In this paper, we refer to models composed exclusively of Linear-MoE layers as pure Linear-MoE models. These models achieve high efficiency during both training and inference, benefiting from the LSM modules embedded in each layer. However, despite these advantages, empirical research~\citep{lieber2024jamba,ren2024samba,waleffe2024empirical} has shown that models relying solely on LSM modules tend to underperform on tasks requiring strong recall capabilities, such as in-context learning (e.g., five-shot MMLU~\citep{hendrycks2020measuring}, Phone-book lookup~\citep{jelassi2024repeat}, Needle In A Haystack~\citep{briakou2023searching}) and long-context reasoning. In such cases, a hybrid architecture that interleaves linear transformer layers with standard transformer layers has proven effective in improving model performance on recall-intensive tasks \cite{yang2024gated,minimax2025minimax,lan2025liger}.

Based on this prior, we propose a hybrid Linear-MoE architecture that combines Linear-MoE layers with standard (MoE) transformer layers. A practical approach for constructing these hybrid models is to periodically substitute certain Linear-MoE layers with standard MoE transformer layers within the model. For instance, in an 4-layer hybrid Linear-MoE model, denoted by "L" for Linear-MoE layers and "N" for normal transformer layers, configurations such as "LLLL" or "LNLN" may be used, depending on the desired ratio of normal transformer layers, which can be adjusted based on user preference.

\begin{figure*}[t]
    \centering
    \vspace{-10mm}
    \includegraphics[width=0.9\textwidth]{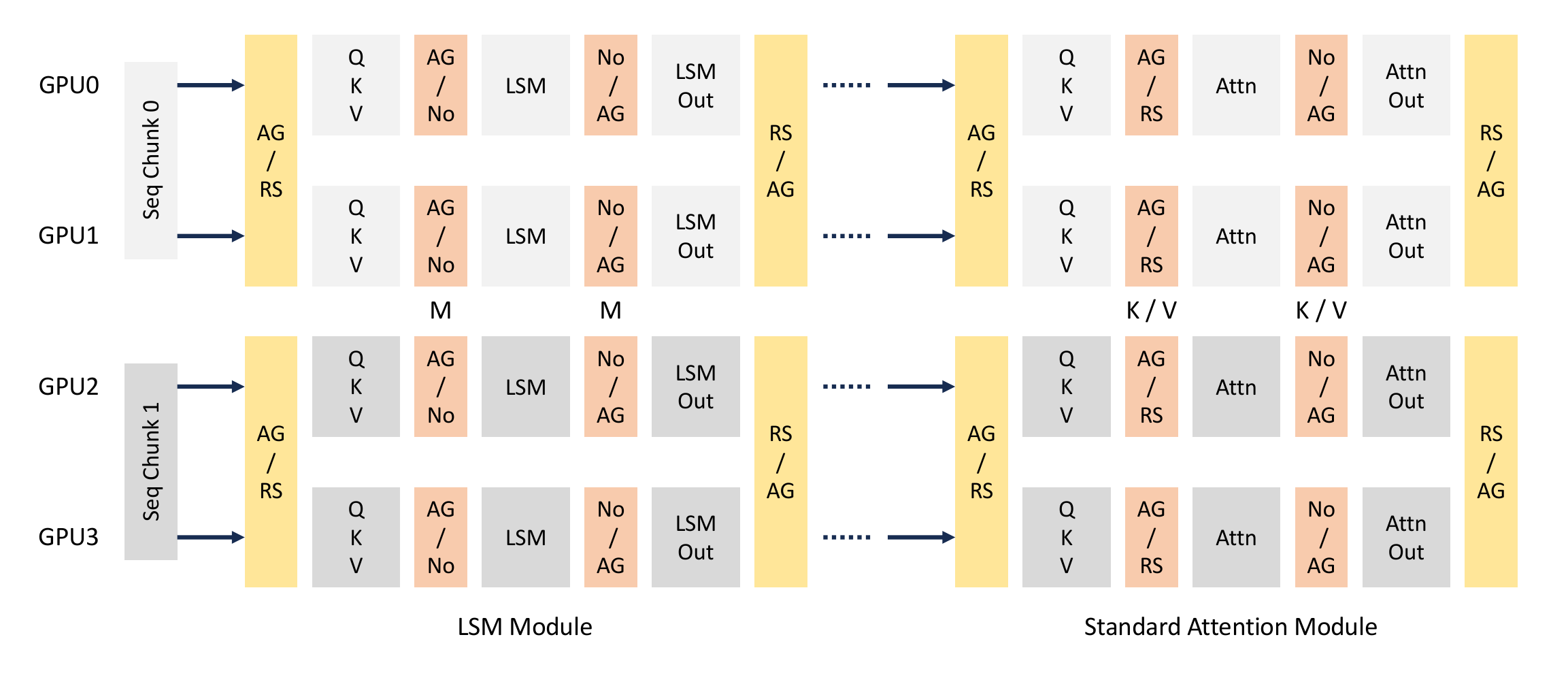}
    \vspace{-4mm}
    \caption{\textbf{Sequence Parallelism Approach on Hybrid Linear-MoE models.} We exemplify the parallelism on the hybrid layers of LSM and standard attention with both TP and SP (both have a dimension of 2). The communication operations colored in yellow and green are for TP and SP, respectively. AG/RS: all-gather in forward and reduce-scatter in backward, RS/AG: reduce-scatter in forward and all-gather in backward, AG/No: all-gather in forward and no-op in backward, No/AG: no-op in forward and all-gather in backward. Note that the SP communication operations for linear attention operate on the memory state $\mathbf{M}_s\in \mathbb{R}^{d\times d}$, while for standard attention, they operate on states $\mathbf{K}_s, \mathbf{V}_s\in \mathbb{R}^{C\times d}$.}
    \vspace{-2mm}
    \label{fig: linear-moe-sp}
\end{figure*}

\subsection{Training}
\label{subsec: training}

\subsubsection{Sequence Parallelism on Linear-MoE}

The existing methods, LASP~\citep{sun2024linear} and its improved version LASP-2~\citep{sun2025lasp}, are designed specifically to leverage the right-product-first property of linear attention techniques for efficient sequence parallelism (SP). LASP employs a point-to-point ring-style communication pattern, facilitating the exchange of incremental memory states across devices. This communication pattern is particularly effective for managing dependencies while minimizing the data transferred between devices, enhancing the scalability of SP. LASP-2 further refines this approach by replacing the ring-style communication with an all-gather collective communication operation, streamlining the entire communication process. This modification not only simplifies the communication structure but also improves the parallelism of computation and communication.

In this work, we extend the capabilities of LASP series to the Linear-MoE system, allowing for the efficient SP training on LSM modules, particularly when dealing with extremely long sequences across large-scale distributed clusters. This extension significantly enhances the scalability and efficiency of training large-scale Linear-MoE models with long-context sequences on extensive compute resources. A detailed breakdown of the SP algorithm on Linear-MoE, with and without masking, is provided in Appendix~\ref{app: sp}.

\subsubsection{Hybrid Model Sequence Parallelism}

Hybrid linear sequence modeling models, which combine linear transformer layers (leveraging LSM methods for token mixing) with standard transformer layers (utilizing conventional self-attention for token mixing), have demonstrated substantial improvements in handling long-context tasks~\citep{lieber2024jamba,ren2024samba,waleffe2024empirical}. This hybrid model is particularly beneficial for tasks with high recall requirements, including five-shot MMLU~\citep{hendrycks2020measuring}, Phone-book lookup~\citep{jelassi2024repeat}, and Needle In A Haystack~\citep{briakou2023searching}, etc.. Our proposed hybrid Linear-MoE models also aim to enhance performance in areas where pure Linear-MoE models have shown limitations, specifically on tasks where recall precision is critical.

Applying SP on pure Linear-MoE models is straightforward, as this form of SP operates exclusively on the LSM modules, leaving the MoE layers unaffected. In hybrid Linear-MoE models, however, implementing SP becomes more complex due to the interleaving of distinct sequence modeling layers. To effectively optimize SP for these hybrid models, we introduce an integrated approach that incorporates SP across both the linear-MoE and standard transformer layers, thus enhancing overall efficiency. We illustrate the approach in Fig.~\ref{fig: linear-moe-sp}, and explain it as below:

\textbf{On LSM Module.} The SP for LSM modules is implemented via a single collective communication operation on the memory state $\mathbf{M}_s \in \mathbb{R}^{d \times d}$. This approach ensures that the communication complexity does not depend on either the sequence or sub-sequence length; rather, it scales only linearly with the SP size $T$, thereby maintaining efficiency in distributed setups.

\textbf{On Standard Attention Module.} 
Context parallelism (CP) is a SP technique used in Megatron-LM that divides input data and activations along the sequence dimension, specifically designed for standard softmax attention. Traditional CP implementations in Megatron-LM rely on a ring-like communication-computation overlap~\citep{liu2023ring}. In contrast, our approach for standard attention modules adopts the all-gather-based strategy used in the pretraining of Llama3~\citep{dubey2024llama}. Rather than utilizing a ring strategy, we perform all-gather communication for $\mathbf K_s$ and $\mathbf V_s$ tensors across devices, followed by local computation of attention output on each device’s chunk of $\mathbf Q_s$. While all-gather communication theoretically has higher latency than ring-based methods, it offers greater flexibility and adaptability for handling different attention masks, such as document-level masks, making it ideal for varying attention patterns. Moreover, the latency of all-gather is minimized since the $\mathbf K_s$ and $\mathbf V_s$ tensors are notably smaller than the $\mathbf Q_s$ tensor, especially with grouped query attention~\citep{ainslie2023gqa}. Consequently, the computational time for generating attention output significantly outweighs the cost of all-gather communication.

\subsubsection{Hybrid Parallelism}
\label{subsec: hybrid parallel}

SP in Linear-MoE allows for a flexible choice of sequence parallel size that can be set to any factor smaller than or divisible by the total number of distributed nodes (i.e., the world size). This flexibility enables splitting input data across both batch and sequence dimensions, creating a combined approach known as data-sequence hybrid parallelism. Standard data parallelism techniques, such as Distributed Data Parallel (DDP)~\citep{li2020pytorch}, can integrate seamlessly with SP in Linear-MoE. Additionally, the sharded data parallelism method, like Distributed Optimizer~\citep{korthikanti2022reducing} in Megatron-Core, is also compatible.

Furthermore, the system provides support for Tensor Parallelism (TP), Pipeline Parallelism (PP), and Expert Parallelism (EP) specifically tailored for Linear-MoE models. In the case of TP, its application to Linear-MoE models is direct and efficient, as detailed in $\S$\ref{subsubsec: TP}. Regarding PP and EP, these parallelism techniques operate on Linear-MoE in much the same way as their original versions since they are not involved in the inner computations of the LSM modules but rather work at the level of complete Linear-MoE blocks or MoE layers. Moreover, TP, PP, and EP can be combined with DP and SP as introduced earlier, enhancing flexibility and scalability for large distributed setups.


\subsubsection{Variable Length}
During pretraining, batches generally consist of sequences with a uniform length. However, in the finetuning phase or during inference, the model often encounters batches containing sequences of different lengths. A common approach to handle this variation is to right-pad each sequence in the batch so that all match the length of the longest sequence in that batch. While straightforward, this padding strategy can lead to inefficiencies, particularly when sequence lengths vary greatly within a batch. For standard transformers, more advanced methods have been introduced to address this issue. These methods include techniques like distributing workloads across GPUs to avoid padding altogether~\citep{zeng2022boosting, zhai2023bytetransformer}, or packing multiple sequences into a single batch while adjusting the attention mask as needed~\citep{ding2024fewer, pouransari2024dataset}. In Linear-MoE, handling variable-length sequences is simplified by processing the entire batch as one continuous long sequence, effectively managing varying sequence lengths without the need for padding.

\begin{figure}[t]
    \centering
    \includegraphics[width=1\columnwidth]{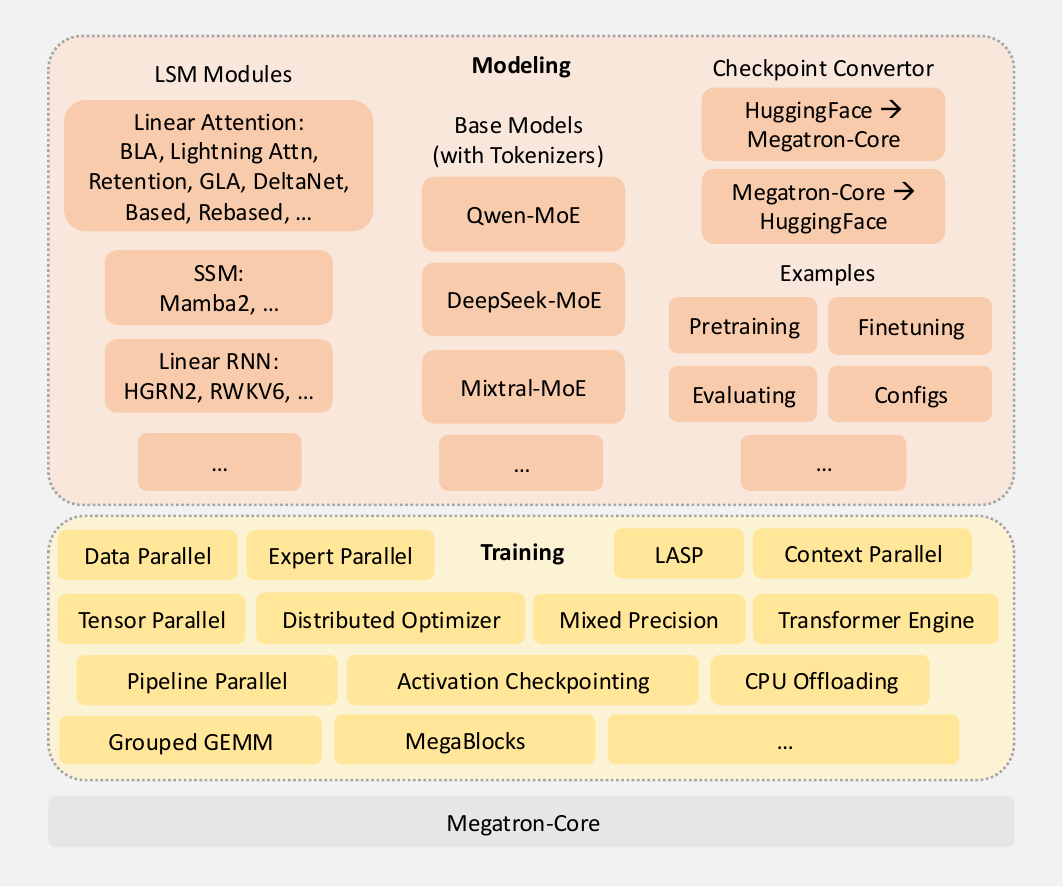}
    \caption{\textbf{Linear-MoE System Implementation.} The Linear-MoE system is composed of two main subsystems: Modeling and Training. It is developed in a non-intrusive manner, utilizing the latest version of Megatron-Core. All components within the system are designed with extensibility in mind, encompassing the LSM modules, base models, examples, and training technologies. This design allows for future enhancements and extensions of the system.}
    \vspace{-2mm}
    \label{fig: system impl}
\end{figure}


\subsection{Implementation}

The implementation of the Linear-MoE system is based on Megatron-Core, an open-source library developed on PyTorch that incorporates optimized GPU techniques and advanced system-level enhancements. As depicted in Fig.~\ref{fig: system impl}, the Linear-MoE system consists of both modeling and training subsystems, facilitating adaptable model building and efficient training specifically for Linear-MoE models. Leveraging the capabilities of Megatron-Core, the Linear-MoE library is fully compatible with all NVIDIA Tensor Core GPUs, including support for FP8 acceleration on NVIDIA Hopper architectures.

The Linear-MoE design approach aims to minimize any invasive changes to Megatron-Core’s source code. Rather than adding new modules directly, Linear-MoE operates independently, allowing users to benefit from the latest LLM practices without disruptions due to updates or changes within Megatron-Core.

\subsubsection{Modeling Subsystem}
Linear-MoE abstracts its LSM modules into modular and composable APIs, providing model developers and researchers with extensive flexibility to design and train large-scale custom Linear-MoE models on accelerated computing platforms. The system includes essential building blocks, such as core components for LSM mechanisms, MoE layers and Linear-MoE blocks, normalization techniques, and embedding methods. To enhance adaptability, LSM mechanisms are organized into three main categories: linear attention, SSM, and linear RNN, with multiple instances available in each. For linear attention, options include basic linear attention (BLA), Lightning Attention, Retention, GLA, DeltaNet, Based, and Rebased; for SSM, we provide Mamba2, the leading SSM model at present; and for linear RNN, options include HGRN2 and RWKV6. As LSM techniques evolve, Linear-MoE will continue to incorporate more LSM methods to ensure users have access to the latest advancements.

Additionally, Linear-MoE offers vital components such as a model library, tokenizers, model converters, usage examples, and a set of supportive toolkits. The model library includes instances of Linear-MoE models that are adapted from state-of-the-art open-source MoE architectures, including Qwen2 MoE, DeepSeekV2 MoE, and Mixtral MoE. These adapted instances are designated as Linear-MoE-Qwen2, Linear-MoE-DeepSeekV2, and Linear-MoE-Mixtral, respectively. These models are implemented following Megatron-Core format, with the standard attention layers replaced by LSM-based token mixing layers, while maintaining the original embedding, normalization, and expert layers unchanged.

\subsubsection{Training Subsystem}

Advanced parallelism techniques, encompassing tensor, sequence, pipeline, context, and MoE expert parallelism, are seamlessly incorporated into the Linear-MoE system through its design on top of the Megatron-Core library. This non-intrusive integration allows Linear-MoE to leverage the robust training capabilities of Megatron-Core, supporting large-scale model training across both standard attention layers and MoE expert layers. However, the inherent parallelism mechanisms, such as TP and SP, were not originally optimized for LSM modules. Additionally, Megatron-Core does not fully support efficient SP for hybrid models containing both LSM modules and standard attention layers. To address these gaps, we elaborate on our TP and SP approaches specifically designed for LSM modules and hybrid models, as discussed in $\S$\ref{subsec: training}.

Further capabilities, including mixed precision, activation recomputation, distributed optimizer, distributed checkpointing, and CPU offloading, are also inherited from Megatron-Core, enhancing model training flexibility and efficiency. And Linear-MoE supports 8-bit floating point (FP8) precision on Hopper GPUs, benefiting from the integration of NVIDIA's Transformer Engine~\citep{micikevicius2022fp8}. This feature optimizes memory usage and accelerates performance during both training and inference stages.

To enhance the training speed of MoE layers, we incorporate MegaBlocks~\citep{megablocks} into our Linear-MoE system. MegaBlocks is designed to optimize MoE training on GPUs by reconfiguring MoE computations using block-sparse operations and developing new block-sparse GPU kernels that effectively manage the inherent dynamism of MoE. In addition, we also integrate the Grouped GEMM library into Linear-MoE, which introduces grouped GEMM kernels in PyTorch, thereby accelerating the computational processes involved in training MoE models.

\subsubsection{Evaluation Module}

In order to facilitate the evaluation on mainstream benchmarks, we have developed offline text generation of Linear-MoE models within the system. Based on this, mature evaluation frameworks such as OpenCompass~\citep{2023opencompass} and LM-Evaluation-Harness~\citep{eval-harness}, are readily available for conducting evaluation tasks on Linear-MoE models. Furthermore, the system facilitates seamless bidirectional conversion between model weights from HuggingFace and Megatron-Core. This functionality enables users to easily leverage pretrained models from HuggingFace for continued pretraining or fine-tuning within the Megatron-Core environment. Additionally, it allows for the assessment of model performance by using HuggingFace's evaluation and inference pipelines on models trained within the Megatron-Core framework.

\begin{table}[t]
\centering
\small
\begin{tabular}{ccccccccc}
\toprule
\textbf{Models} & \textbf{A0.3B-2B} & \textbf{A1B-7B} \\
\midrule
Hidden Dimension & 1024 & 2048 \\
FFN Dimension & 896 & 1024 \\
Num of Heads & 8 & 16 \\
Num of Layers & 12 & 16 \\
Num of Act Experts & 8 & 8\\
Num of Experts & 64 & 64 \\
LR & 1e-4 & 1e-5 \\
Minimum LR & 1e-5 & 1e-6 \\
LR Scheduler & Cosine & Cosine \\
Seq Length & 2048 & 2048 \\
Training Tokens & 15B & 100B \\
\bottomrule
\end{tabular}
\addtolength{\tabcolsep}{2.5pt}    
\centering
\caption{\textbf{Linear-MoE Family Models and Training Configurations.}  A0.3B-2B indicates that the Linear-MoE model has a total of 2 billion parameters, with 0.3 billion parameters activated. The same for A1B-7B.}
\label{tab:model_config}
\end{table}

\begin{figure}[t]
    \centering
    \includegraphics[width=\columnwidth]{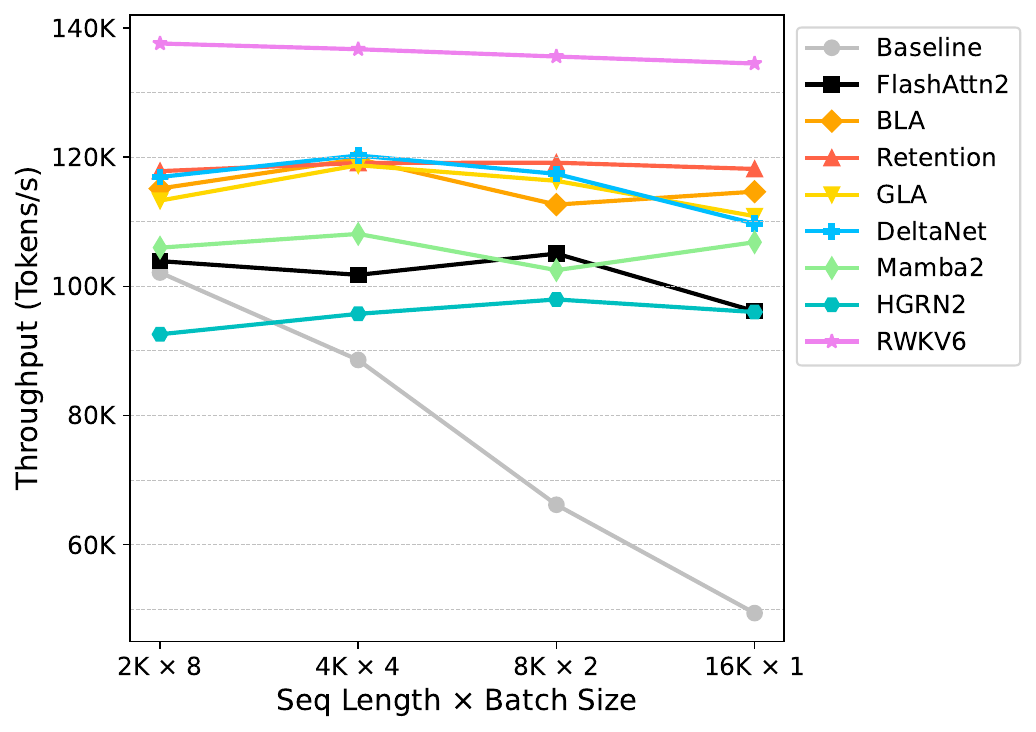}
    \vspace{-4mm}
    \caption{\textbf{Training Throughput (Tokens/s).} As sequence length increases, the throughput of Baseline declines significantly, whereas LSM models maintain stable training efficiency.}
    \vspace{-2mm}
    \label{fig: throughput}
\end{figure}

\begin{table*}[t]
\centering
\small
\begin{tabular}{ccccccccc}
\toprule
\multirow{2}{*}{\textbf{\makecell[c]{Seq Length $\times$ Batch Size}}} & \multicolumn{2}{c}{\textbf{2K $\times$ 8}} & \multicolumn{2}{c}{\textbf{4K $\times$ 4}} & \multicolumn{2}{c}{\textbf{8K $\times$ 2}} & \multicolumn{2}{c}{\textbf{16K $\times$ 1}} \\
\cmidrule(lr){2-3} \cmidrule(lr){4-5} \cmidrule(lr){6-7} \cmidrule(lr){8-9}
 & Mem. & Thpt. & Mem. & Thpt. & Mem. & Thpt. & Mem. & Thpt. \\
\midrule
\textbf{Baseline} & 40.74  & 102.14 & 41.42 & 88.60 & 42.93 & 66.17 & 47.08 & 49.39 \\
\textbf{FlashAttn-2} & 38.96 & 103.91 & 39.10 & 101.78 & 39.57 & 105.08 & 41.51 & 96.16 \\
\textbf{Basic LA} & 42.69 & 115.16 & 43.85 & 119.72 & 42.71 & 112.66 & 43.00 & 114.67 \\
\textbf{Retention} & 42.71 & 117.85 & 42.66 & 119.11 & 42.73 & 119.16 & 42.65 & 118.19 \\
\textbf{GLA} & 43.87 & 113.29 & 43.73 & 118.77 & 43.63 & 116.34 & 43.60 & 110.87 \\
\textbf{DeltaNet} & 43.33 & 116.95 & 43.34 & 120.27 & 43.31 & 117.43 & 43.32 & 109.72 \\
\textbf{Mamba2} & 45.63 & 105.99 & 45.94 & 108.13 & 47.16 & 102.51 & 44.97 & 106.84 \\
\textbf{HGRN2} & 46.03 & 92.58 & 46.14 & 95.74 & 45.56 & 97.98 & 44.97 & 96.02 \\
\textbf{RWKV6} & 47.11 & 137.62 & 47.12 & 136.73 & 47.11 & 135.60 & 47.12 & 134.51 \\
\bottomrule
\end{tabular}
\addtolength{\tabcolsep}{2.5pt}    
\centering
\caption{\textbf{Quantitative Training Efficiency Results.} We experiment on 8 A100 GPUs and report the max allocated GPU memory (GB) and throughput ($\times 10^3$ tokens/s) of A0.3B-2B model instances with varying input sequence lengthes and batch sizes.}
\label{tab:efficiency_results}
\end{table*}

\section{Empirical Study}


\subsection{Experiment Setup} 

\textbf{Models and Dataset.}
We conduct experiments on two Linear-MoE model series: A0.3B-2B and A1B-7B. A0.3B-2B denotes a Linear-MoE model containing a total of 2 billion parameters, with 0.3 billion parameters activated. The same applies for the A1B-7B model. Each series consists of several model instances, each incorporating a distinct instance of the LSM module. The specific LSM module instances used in our experiments include: basic linear attention (BLA)~\citep{katharopoulos2020transformers}, Retentive Network (Retention)~\citep{sun2023retentive}, Gated Linear Attention (GLA)~\citep{yang2023gated}, DeltaNet~\citep{Schlag2021LinearTA}, Mamba2~\citep{dao2024transformers}, HGRN2~\citep{qin2024hgrn2}, and RWKV6~\citep{peng-etal-2023-rwkv,peng2024eagle}, all implemented in Triton. These model instances are evaluated against models with standard attention implementation in Megatron-Core (referred to as Baseline) and the FlashAttention-2~\citep{dao2023flashattention} implemented in Transformer Engine (in CUDA). 

To implement the Linear-MoE model instances, we utilize the Qwen2 MoE architecture~\citep{qwen2} as the base model. All models are pretrained from scratch on a portion of the SlimPajama dataset~\citep{cerebras2023slimpajama}. This dataset originally contains 627 billion tokens, we restrict our experiments to the first two chunks of the dataset, totaling approximately 100 billion tokens. The Qwen2 tokenizer is employed throughout the training processes.

\textbf{Training Configurations.} Table~\ref{tab:model_config} details the training configurations for both Linear-MoE model series. We employ the Adam optimizer~\citep{kingma2014adam} along with parallelism techniques, including TP and EP. Each pretraining run is performed on a node with eight A100 80G GPUs.

\subsection{Training and Inference Efficiency}

\begin{figure*}[t]
    \centering
    \includegraphics[width=0.8\linewidth]{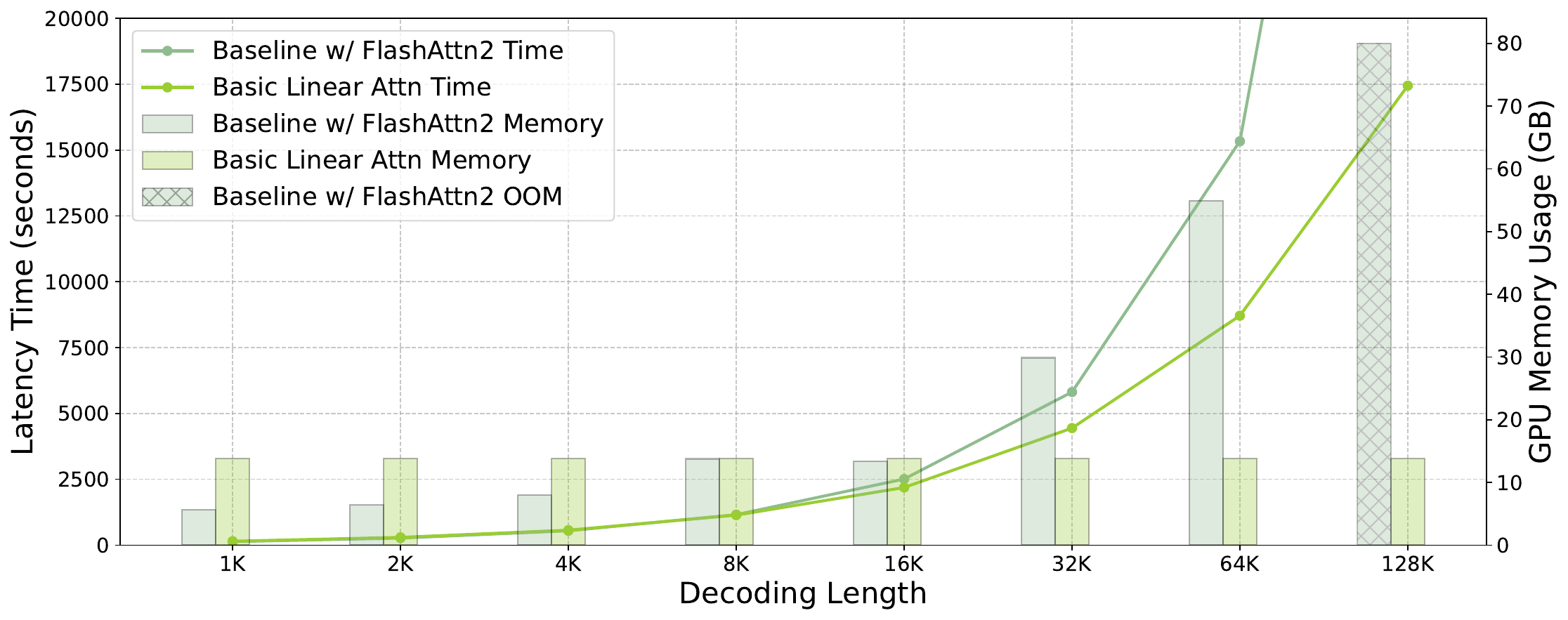}
    \vspace{-2mm}
    \caption{\textbf{Inference Efficiency of A0.3B-2B Model Instances.} We variate the decoding length from 1K to 128K with fixed batch size of 16 on single A800 80GB GPU to evaluate the Baseline w/ FlashAttention-2 and the Linear-MoE w/ Basic Linear Attention in terms of inference latency time and GPU memory usage.}
    \vspace{-2mm}
    \label{fig: inference speed}
\end{figure*}

\begin{table}[t]
\centering
\small
\begin{tabular}{ccc}
\toprule
\textbf{MoE Optimization} & \textbf{Memory (GB)} & \textbf{Time/Iter (ms)}  \\
\midrule
Baseline & 35.28 &  1565.6  \\
Grouped GEMM & 35.01 & 455.4 \\
MegaBlocks & 36.90 & 348.8 \\
\bottomrule
\end{tabular}
\vspace{4mm}

\begin{tabular}{ccc|cc}
\toprule
\textbf{EP} & \textbf{TP} & \textbf{PP} & \textbf{Memory (GB)} & \textbf{Time/Iter (ms)} \\
\midrule
1 & 1 & 1 & 35.28 & 1565.6 \\
8 & 1 & 1 & 22.98 &  739.4 \\
1 & 8 & 1 & 10.04 & 6879.0 \\
1 & 1 & 8 & 8.89 & 1820.2  \\
2 & 2 & 2 & 12.90 & 1684.9 \\
\bottomrule
\end{tabular}
\addtolength{\tabcolsep}{2.5pt}    
\centering
\caption{\textbf{Above: MoE Optimization.} \textbf{Below: Distributed training efficiency under different parallelism settings.} We report the memory usage per GPU (GB) and elapsed time per iteration (ms) while training the A0.3B-2B model with a sequence length of 2048 and a batch size of 4, using a node equipped with 8 A100 GPUs. The Baseline refers to the MoE implementation in Megatron-Core, which is used without any optimizations.}
\label{tab:megablock and parallelism}
\vspace{-2mm}
\end{table}

\begin{figure*}[t]
\centering
\includegraphics[width=0.85\columnwidth]{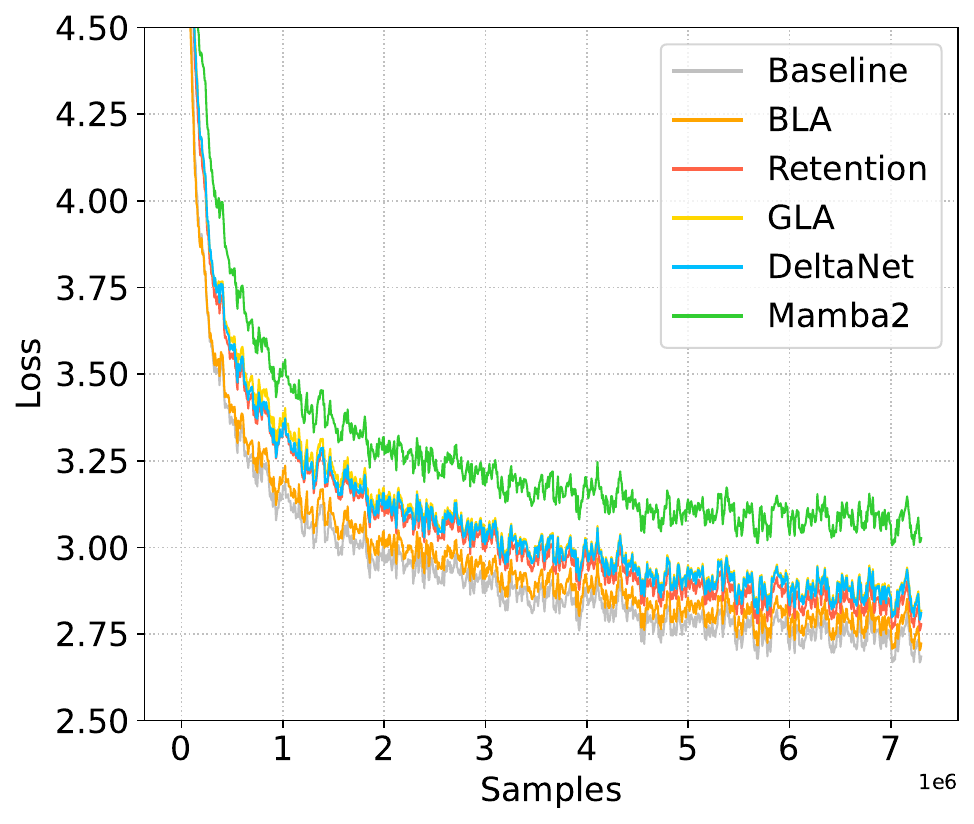}
\includegraphics[width=0.85\columnwidth]{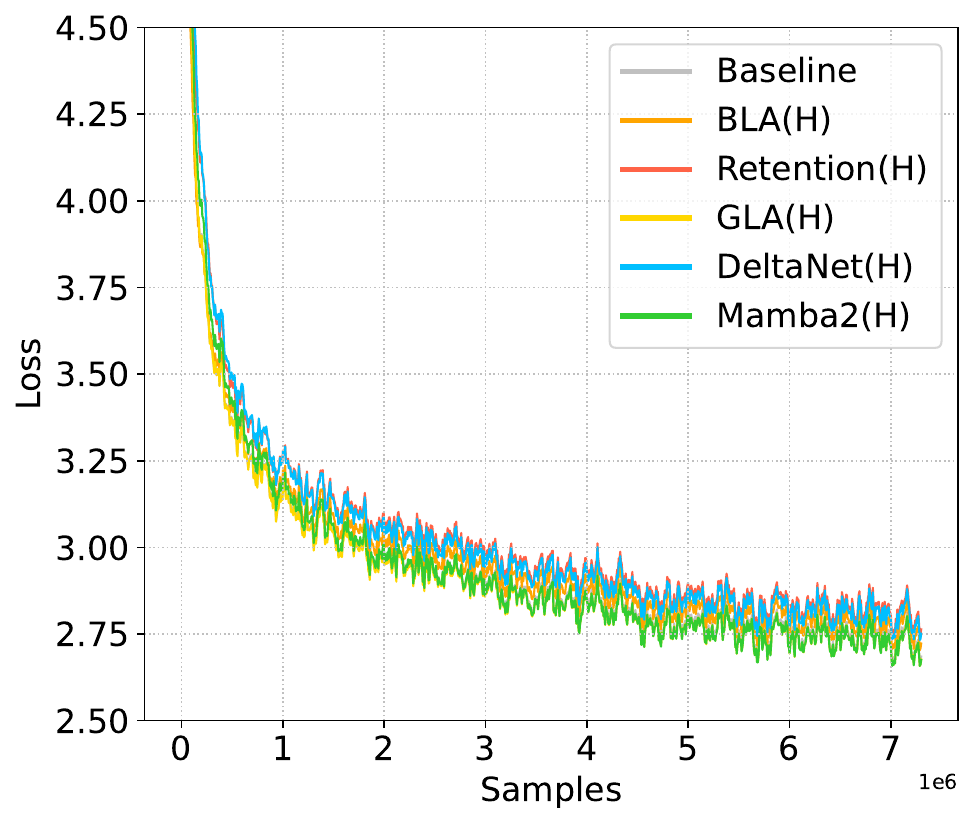}
\vspace{-2mm}
\caption{\textbf{Training Loss Curves of A0.3B-2B Model Instances.} Left: pure Linear-MoE models; Right: hybrid Linear-MoE models. Linear-MoE shows competitive training convergence performance compared to the standard attention Baseline.}
\label{fig: loss}
\vspace{-2mm}
\end{figure*}

We perform experiments to evaluate the training efficiency of the Linear-MoE system, focusing on throughput and GPU memory requirements using eight A100 GPUs. For training the sparse MoE models, we set the EP size to 8. During the experiments, we maintain a total of 16K input tokens per iteration, while varying the input sequence lengths across \{2K, 4K, 8K, 16K\} with corresponding batch sizes of \{8, 4, 2, 1\}. As illustrated in Table \ref{tab:efficiency_results} and Fig. \ref{fig: throughput}, we observe that the standard attention Baseline shows a significant quadratic increase in memory usage and a decline in throughput as the input sequence lengths grow. FlashAttention-2 also demonstrates notable variations in both memory footprint and throughput, when the sequence length reaches 16K. In contrast, the Linear-MoE models, which incorporate LSM, exhibit relatively stable GPU memory consumption and consistent throughput when the sequence length increases, but number of input tokens remains fixed.

We also perform experiments to compare the inference efficiency of Linear-MoE (using Basic LA) with the Baseline (using FlashAttention-2). The results, shown in Table~\ref{fig: inference speed}, reveal that Linear-MoE offers a significant speed advantage when the decoding length exceeds 16K. Additionally, its memory usage remains constant, which is a key benefit resulting from the adoption of LSM.

Furthermore, to highlight the efficiency benefits of the Linear-MoE training subsystem, we conduct ablation studies on MoE optimization techniques and parallelism training methods. The results of these experiments are presented in Table \ref{tab:megablock and parallelism}. It is evident that the implementation of MoE optimization techniques, specifically Grouped GEMM and MegaBlocks, significantly reduces the elapsed time for each iteration. Additionally, the various parallelism training techniques each demonstrate their own advantages in terms of memory footprint and overall training efficiency.


\subsection{Training Loss and Evaluation}

To evaluate the overall training performance of the Linear-MoE models, we pretrain the A0.3B-2B and A1B-7B model instances using 15B and 100B tokens, respectively. We test both pure and hybrid model configurations; for the hybrid models, we incorporate one quarter of standard transformer MoE layers throughout the architecture. For instance, in the 12-layer A0.3B-2B model, the hybrid configuration follows the pattern "LLLNLLLNLLLN", while the 16-layer A1B-7B model adopts the pattern "LLLNLLLNLLLNLLLN".

The training loss curves for the A0.3B-2B model instances, which include both pure and hybrid Linear-MoE models, are presented in Fig.~\ref{fig: loss}. The results demonstrate that the pure Linear-MoE architecture achieves competitive convergence performance compared to the standard attention Baseline. Moreover, the hybrid models exhibit more stable convergence and consistent performance when compared with the Baseline. Additional experiment results such as benchmark evaluations and training loss curves of A1B-7B models can be found in Appendix~\ref{app: additional exp}. Both the A0.3B-2B and A1B-7B Linear-MoE model series show competitive performance on various benchmarks, and it is verified that hybrid models usually perform better than the pure linear models.

\section{Conclusion}
In this paper, we introduced Linear-MoE, a novel product-level system designed to integrate LSM with MoE, aiming to advance both the efficiency and scalability of existing large models. By combining linear-complexity sequence modeling capabilities of LSM with sparsely activated MoE layers, Linear-MoE achieves high performance while addressing computational and memory constraints common in large model training and deployment. The dual subsystems: Modeling and Training, provide a flexible and extensible framework that supports diverse LSM methods and advanced parallelism techniques, including specific sequence parallelism for handling long input sequences efficiently. We also explored hybrid models that further enhance adaptability by incorporating standard Transformer layers. Our experimental results demonstrate that Linear-MoE achieves significant efficiency gains while maintaining strong performance across various benchmarks. These findings highlight the potential of Linear-MoE as the next generation of foundation model architecture.

\section*{Limitations}

Despite the promising results demonstrated in this paper, there are several limitations to the Linear-MoE framework. First, while the system successfully combines LSM with MoE to enhance efficiency, the integration of different LSM methods and MoE layers may introduce complexity in hyperparameter tuning, which could impact model performance under certain configurations. Additionally, the scalability of Linear-MoE in extremely large-scale settings, such as beyond the model sizes tested in our experiments (A0.3B-2B and A1B-7B), remains an area for further investigation. Moreover, while the system supports various parallelism techniques, their effectiveness on diverse hardware architectures, particularly in resource-constrained environments, needs more comprehensive evaluation. Therefore, future work should focus on further optimizing the system for a broader set of use cases and exploring additional hybrid modeling strategies.



\bibliography{custom}

\begin{thebibliography}{78}
\providecommand{\natexlab}[1]{#1}

\bibitem[{Ainslie et~al.(2023)Ainslie, Lee-Thorp, de~Jong, Zemlyanskiy, Lebr{\'o}n, and Sanghai}]{ainslie2023gqa}
Joshua Ainslie, James Lee-Thorp, Michiel de~Jong, Yury Zemlyanskiy, Federico Lebr{\'o}n, and Sumit Sanghai. 2023.
\newblock {GQA}: Training generalized multi-query transformer models from multi-head checkpoints.
\newblock \emph{arXiv preprint arXiv:2305.13245}.

\bibitem[{Behrouz et~al.(2024)Behrouz, Zhong, and Mirrokni}]{behrouz2024titans}
Ali Behrouz, Peilin Zhong, and Vahab Mirrokni. 2024.
\newblock Titans: Learning to memorize at test time.
\newblock \emph{arXiv preprint arXiv:2501.00663}.

\bibitem[{Bisk et~al.(2020)Bisk, Zellers, Bras, Gao, and Choi}]{Bisk2020piqa}
Yonatan Bisk, Rowan Zellers, Ronan~Le Bras, Jianfeng Gao, and Yejin Choi. 2020.
\newblock Piqa: Reasoning about physical commonsense in natural language.
\newblock In \emph{Thirty-Fourth AAAI Conference on Artificial Intelligence}.

\bibitem[{Briakou et~al.(2023)Briakou, Cherry, and Foster}]{briakou2023searching}
Eleftheria Briakou, Colin Cherry, and George Foster. 2023.
\newblock Searching for needles in a haystack: On the role of incidental bilingualism in palm's translation capability.
\newblock \emph{arXiv preprint arXiv:2305.10266}.

\bibitem[{Cai et~al.(2024)Cai, Jiang, Wang, Tang, Kim, and Huang}]{cai2024survey}
Weilin Cai, Juyong Jiang, Fan Wang, Jing Tang, Sunghun Kim, and Jiayi Huang. 2024.
\newblock A survey on mixture of experts.
\newblock \emph{arXiv preprint arXiv:2407.06204}.

\bibitem[{Chintala(2023)}]{gpt4}
Soumith Chintala. 2023.
\newblock \href {https://x.com/soumithchintala/status/1671267150101721090} {Gpt-4 moe}.

\bibitem[{Clark et~al.(2018)Clark, Cowhey, Etzioni, Khot, Sabharwal, Schoenick, and Tafjord}]{clark2018think}
Peter Clark, Isaac Cowhey, Oren Etzioni, Tushar Khot, Ashish Sabharwal, Carissa Schoenick, and Oyvind Tafjord. 2018.
\newblock \href {https://arxiv.org/abs/1803.05457} {Think you have solved question answering? try arc, the ai2 reasoning challenge}.
\newblock \emph{Preprint}, arXiv:1803.05457.

\bibitem[{Contributors(2023)}]{2023opencompass}
OpenCompass Contributors. 2023.
\newblock Opencompass: A universal evaluation platform for foundation models.
\newblock \url{https://github.com/open-compass/opencompass}.

\bibitem[{Dao(2023)}]{dao2023flashattention}
Tri Dao. 2023.
\newblock Flashattention-2: Faster attention with better parallelism and work partitioning.
\newblock \emph{arXiv preprint arXiv:2307.08691}.

\bibitem[{Dao and Gu(2024)}]{dao2024transformers}
Tri Dao and Albert Gu. 2024.
\newblock Transformers are ssms: Generalized models and efficient algorithms through structured state space duality.
\newblock \emph{arXiv preprint arXiv:2405.21060}.

\bibitem[{Ding et~al.(2024)Ding, Wang, Paolini, Kumar, Deoras, Roth, and Soatto}]{ding2024fewer}
Hantian Ding, Zijian Wang, Giovanni Paolini, Varun Kumar, Anoop Deoras, Dan Roth, and Stefano Soatto. 2024.
\newblock Fewer truncations improve language modeling.
\newblock \emph{arXiv preprint arXiv:2404.10830}.

\bibitem[{Du et~al.(2025)Du, Sun, Lan, Hu, and Cheng}]{du2025mom}
Jusen Du, Weigao Sun, Disen Lan, Jiaxi Hu, and Yu~Cheng. 2025.
\newblock \href {https://arxiv.org/abs/2502.13685} {Mom: Linear sequence modeling with mixture-of-memories}.
\newblock \emph{Preprint}, arXiv:2502.13685.

\bibitem[{Dubey et~al.(2024)Dubey, Jauhri, Pandey, Kadian, Al-Dahle, Letman, Mathur, Schelten, Yang, Fan et~al.}]{dubey2024llama}
Abhimanyu Dubey, Abhinav Jauhri, Abhinav Pandey, Abhishek Kadian, Ahmad Al-Dahle, Aiesha Letman, Akhil Mathur, Alan Schelten, Amy Yang, Angela Fan, et~al. 2024.
\newblock The llama 3 herd of models.
\newblock \emph{arXiv preprint arXiv:2407.21783}.

\bibitem[{Fedus et~al.(2022)Fedus, Zoph, and Shazeer}]{fedus2022switch}
William Fedus, Barret Zoph, and Noam Shazeer. 2022.
\newblock Switch transformers: Scaling to trillion parameter models with simple and efficient sparsity.
\newblock \emph{Journal of Machine Learning Research}, 23(120):1--39.

\bibitem[{Gale et~al.(2023)Gale, Narayanan, Young, and Zaharia}]{megablocks}
Trevor Gale, Deepak Narayanan, Cliff Young, and Matei Zaharia. 2023.
\newblock {MegaBlocks: Efficient Sparse Training with Mixture-of-Experts}.
\newblock \emph{Proceedings of Machine Learning and Systems}, 5.

\bibitem[{Gao et~al.(2023)Gao, Tow, Abbasi, Biderman, Black, DiPofi, Foster, Golding, Hsu, Le~Noac'h, Li, McDonell, Muennighoff, Ociepa, Phang, Reynolds, Schoelkopf, Skowron, Sutawika, Tang, Thite, Wang, Wang, and Zou}]{eval-harness}
Leo Gao, Jonathan Tow, Baber Abbasi, Stella Biderman, Sid Black, Anthony DiPofi, Charles Foster, Laurence Golding, Jeffrey Hsu, Alain Le~Noac'h, Haonan Li, Kyle McDonell, Niklas Muennighoff, Chris Ociepa, Jason Phang, Laria Reynolds, Hailey Schoelkopf, Aviya Skowron, Lintang Sutawika, Eric Tang, Anish Thite, Ben Wang, Kevin Wang, and Andy Zou. 2023.
\newblock \href {https://doi.org/10.5281/zenodo.10256836} {A framework for few-shot language model evaluation}.

\bibitem[{Gu and Dao(2023)}]{gu2023mamba}
Albert Gu and Tri Dao. 2023.
\newblock Mamba: Linear-time sequence modeling with selective state spaces.
\newblock \emph{arXiv preprint arXiv:2312.00752}.

\bibitem[{Gu et~al.(2020)Gu, Dao, Ermon, Rudra, and R{\'e}}]{gu2020hippo}
Albert Gu, Tri Dao, Stefano Ermon, Atri Rudra, and Christopher R{\'e}. 2020.
\newblock Hippo: Recurrent memory with optimal polynomial projections.
\newblock \emph{Advances in neural information processing systems}, 33:1474--1487.

\bibitem[{Gu et~al.(2022{\natexlab{a}})Gu, Goel, Gupta, and R{\'e}}]{gu2022parameterization}
Albert Gu, Karan Goel, Ankit Gupta, and Christopher R{\'e}. 2022{\natexlab{a}}.
\newblock On the parameterization and initialization of diagonal state space models.
\newblock \emph{Advances in Neural Information Processing Systems}, 35:35971--35983.

\bibitem[{Gu et~al.(2022{\natexlab{b}})Gu, Goel, and R\'e}]{gu2022efficiently}
Albert Gu, Karan Goel, and Christopher R\'e. 2022{\natexlab{b}}.
\newblock Efficiently modeling long sequences with structured state spaces.
\newblock In \emph{The International Conference on Learning Representations ({ICLR})}.

\bibitem[{Gupta et~al.(2022)Gupta, Gu, and Berant}]{gupta2022diagonal}
Ankit Gupta, Albert Gu, and Jonathan Berant. 2022.
\newblock Diagonal state spaces are as effective as structured state spaces.
\newblock \emph{Advances in Neural Information Processing Systems}, 35:22982--22994.

\bibitem[{Hendrycks et~al.(2020)Hendrycks, Burns, Basart, Zou, Mazeika, Song, and Steinhardt}]{hendrycks2020measuring}
Dan Hendrycks, Collin Burns, Steven Basart, Andy Zou, Mantas Mazeika, Dawn Song, and Jacob Steinhardt. 2020.
\newblock Measuring massive multitask language understanding.
\newblock \emph{arXiv preprint arXiv:2009.03300}.

\bibitem[{Hu et~al.(2024)Hu, Lan, Zhou, Wen, and Liang}]{hu2024timessm}
Jiaxi Hu, Disen Lan, Ziyu Zhou, Qingsong Wen, and Yuxuan Liang. 2024.
\newblock \href {https://arxiv.org/abs/2405.16312} {Time-ssm: Simplifying and unifying state space models for time series forecasting}.
\newblock \emph{Preprint}, arXiv:2405.16312.

\bibitem[{Jacobs et~al.(1991)Jacobs, Jordan, Nowlan, and Hinton}]{Jacobs_Jordan_Nowlan_Hinton_1991}
Robert~A. Jacobs, Michael~I. Jordan, Steven~J. Nowlan, and Geoffrey~E. Hinton. 1991.
\newblock \href {https://doi.org/10.1162/neco.1991.3.1.79} {Adaptive mixtures of local experts}.
\newblock \emph{Neural Computation}, page 79–87.

\bibitem[{Jelassi et~al.(2024)Jelassi, Brandfonbrener, Kakade, and Malach}]{jelassi2024repeat}
Samy Jelassi, David Brandfonbrener, Sham~M Kakade, and Eran Malach. 2024.
\newblock Repeat after me: Transformers are better than state space models at copying.
\newblock \emph{arXiv preprint arXiv:2402.01032}.

\bibitem[{Jiang et~al.(2024)Jiang, Sablayrolles, Roux, Mensch, Savary, Bamford, Chaplot, Casas, Hanna, Bressand et~al.}]{jiang2024mixtral}
Albert~Q Jiang, Alexandre Sablayrolles, Antoine Roux, Arthur Mensch, Blanche Savary, Chris Bamford, Devendra~Singh Chaplot, Diego de~las Casas, Emma~Bou Hanna, Florian Bressand, et~al. 2024.
\newblock Mixtral of experts.
\newblock \emph{arXiv preprint arXiv:2401.04088}.

\bibitem[{Katharopoulos et~al.(2020)Katharopoulos, Vyas, Pappas, and Fleuret}]{katharopoulos2020transformers}
Angelos Katharopoulos, Apoorv Vyas, Nikolaos Pappas, and Fran{\c{c}}ois Fleuret. 2020.
\newblock Transformers are rnns: Fast autoregressive transformers with linear attention.
\newblock In \emph{International Conference on Machine Learning}, pages 5156--5165. PMLR.

\bibitem[{Kingma and Ba(2014)}]{kingma2014adam}
Diederik~P Kingma and Jimmy Ba. 2014.
\newblock Adam: A method for stochastic optimization.
\newblock \emph{arXiv preprint arXiv:1412.6980}.

\bibitem[{Korthikanti et~al.(2022)Korthikanti, Casper, Lym, McAfee, Andersch, Shoeybi, and Catanzaro}]{korthikanti2022reducing}
Vijay Korthikanti, Jared Casper, Sangkug Lym, Lawrence McAfee, Michael Andersch, Mohammad Shoeybi, and Bryan Catanzaro. 2022.
\newblock \href {https://arxiv.org/abs/2205.05198} {Reducing activation recomputation in large transformer models}.
\newblock \emph{Preprint}, arXiv:2205.05198.

\bibitem[{Lan et~al.(2025)Lan, Sun, Hu, Du, and Cheng}]{lan2025liger}
Disen Lan, Weigao Sun, Jiaxi Hu, Jusen Du, and Yu~Cheng. 2025.
\newblock \href {https://arxiv.org/abs/2503.01496} {Liger: Linearizing large language models to gated recurrent structures}.
\newblock \emph{Preprint}, arXiv:2503.01496.

\bibitem[{Lepikhin et~al.(2020)Lepikhin, Lee, Xu, Chen, Firat, Huang, Krikun, Shazeer, and Chen}]{lepikhin2020gshard}
Dmitry Lepikhin, HyoukJoong Lee, Yuanzhong Xu, Dehao Chen, Orhan Firat, Yanping Huang, Maxim Krikun, Noam Shazeer, and Zhifeng Chen. 2020.
\newblock Gshard: Scaling giant models with conditional computation and automatic sharding.
\newblock \emph{arXiv preprint arXiv:2006.16668}.

\bibitem[{Lewis et~al.(2021)Lewis, Bhosale, Dettmers, Goyal, and Zettlemoyer}]{lewis2021base}
Mike Lewis, Shruti Bhosale, Tim Dettmers, Naman Goyal, and Luke Zettlemoyer. 2021.
\newblock Base layers: Simplifying training of large, sparse models.
\newblock In \emph{International Conference on Machine Learning}, pages 6265--6274. PMLR.

\bibitem[{Li et~al.(2023)Li, Zhang, Koto, Yang, Zhao, Gong, Duan, and Baldwin}]{li2023cmmlu}
Haonan Li, Yixuan Zhang, Fajri Koto, Yifei Yang, Hai Zhao, Yeyun Gong, Nan Duan, and Timothy Baldwin. 2023.
\newblock \href {https://arxiv.org/abs/2306.09212} {Cmmlu: Measuring massive multitask language understanding in chinese}.
\newblock \emph{Preprint}, arXiv:2306.09212.

\bibitem[{Li et~al.(2020)Li, Zhao, Varma, Salpekar, Noordhuis, Li, Paszke, Smith, Vaughan, Damania, and Chintala}]{li2020pytorch}
Shen Li, Yanli Zhao, Rohan Varma, Omkar Salpekar, Pieter Noordhuis, Teng Li, Adam Paszke, Jeff Smith, Brian Vaughan, Pritam Damania, and Soumith Chintala. 2020.
\newblock \href {https://arxiv.org/abs/2006.15704} {Pytorch distributed: Experiences on accelerating data parallel training}.
\newblock \emph{Preprint}, arXiv:2006.15704.

\bibitem[{Lieber et~al.(2024)Lieber, Lenz, Bata, Cohen, Osin, Dalmedigos, Safahi, Meirom, Belinkov, Shalev-Shwartz et~al.}]{lieber2024jamba}
Opher Lieber, Barak Lenz, Hofit Bata, Gal Cohen, Jhonathan Osin, Itay Dalmedigos, Erez Safahi, Shaked Meirom, Yonatan Belinkov, Shai Shalev-Shwartz, et~al. 2024.
\newblock Jamba: A hybrid transformer-mamba language model.
\newblock \emph{arXiv preprint arXiv:2403.19887}.

\bibitem[{Liu et~al.(2024)Liu, Feng, Wang, Wang, Liu, Zhao, Dengr, Ruan, Dai, Guo et~al.}]{liu2024deepseek}
Aixin Liu, Bei Feng, Bin Wang, Bingxuan Wang, Bo~Liu, Chenggang Zhao, Chengqi Dengr, Chong Ruan, Damai Dai, Daya Guo, et~al. 2024.
\newblock Deepseek-v2: A strong, economical, and efficient mixture-of-experts language model.
\newblock \emph{arXiv preprint arXiv:2405.04434}.

\bibitem[{Liu et~al.(2023)Liu, Zaharia, and Abbeel}]{liu2023ring}
Hao Liu, Matei Zaharia, and Pieter Abbeel. 2023.
\newblock \href {https://arxiv.org/abs/2310.01889} {Ring attention with blockwise transformers for near-infinite context}.
\newblock \emph{Preprint}, arXiv:2310.01889.

\bibitem[{Lu et~al.(2025)Lu, Fan, Wei, Qu, Chen, and Cheng}]{lu2025twin}
Zhenyi Lu, Chenghao Fan, Wei Wei, Xiaoye Qu, Dangyang Chen, and Yu~Cheng. 2025.
\newblock Twin-merging: Dynamic integration of modular expertise in model merging.
\newblock \emph{Advances in Neural Information Processing Systems}, 37:78905--78935.

\bibitem[{Micikevicius et~al.(2022)Micikevicius, Stosic, Burgess, Cornea, Dubey, Grisenthwaite, Ha, Heinecke, Judd, Kamalu et~al.}]{micikevicius2022fp8}
Paulius Micikevicius, Dusan Stosic, Neil Burgess, Marius Cornea, Pradeep Dubey, Richard Grisenthwaite, Sangwon Ha, Alexander Heinecke, Patrick Judd, John Kamalu, et~al. 2022.
\newblock Fp8 formats for deep learning.
\newblock \emph{arXiv preprint arXiv:2209.05433}.

\bibitem[{MiniMax et~al.(2025)MiniMax, Li, Gong, Yang, Shan, Liu, Zhu, Zhang, Guo, Chen, Li, Jiao, Li, Zhang, Sun, Dong, Zhu, Zhuang, Song, Zhu, Han, Li, Xie, Xu, Yan, Zhang, Xiao, Kang, Han, Wang, Yu, Feng, Zheng, Chai, Xing, Ju, Chi, Zhang, Huang, Niu, Li, Zhao, Yang, Xu, Wang, Wang, Li, Leng, Shi, Yu, Li, Zhu, Huang, Liang, Sun, Sun, Cheng, Li, Song, Su, Han, Zhang, Hou, Min, Zou, Shen, Gong, Zhu, Zhou, Zhong, Hu, Fan, Yu, Yang, Li, Huang, Li, Huang, Xu, Mao, Li, Li, Tao, Ying, Cong, Qin, Fan, Yu, Jiang, and Wu}]{minimax2025minimax}
MiniMax, Aonian Li, Bangwei Gong, Bo~Yang, Boji Shan, Chang Liu, Cheng Zhu, Chunhao Zhang, Congchao Guo, Da~Chen, Dong Li, Enwei Jiao, Gengxin Li, Guojun Zhang, Haohai Sun, Houze Dong, Jiadai Zhu, Jiaqi Zhuang, Jiayuan Song, Jin Zhu, Jingtao Han, Jingyang Li, Junbin Xie, Junhao Xu, Junjie Yan, Kaishun Zhang, Kecheng Xiao, Kexi Kang, Le~Han, Leyang Wang, Lianfei Yu, Liheng Feng, Lin Zheng, Linbo Chai, Long Xing, Meizhi Ju, Mingyuan Chi, Mozhi Zhang, Peikai Huang, Pengcheng Niu, Pengfei Li, Pengyu Zhao, Qi~Yang, Qidi Xu, Qiexiang Wang, Qin Wang, Qiuhui Li, Ruitao Leng, Shengmin Shi, Shuqi Yu, Sichen Li, Songquan Zhu, Tao Huang, Tianrun Liang, Weigao Sun, Weixuan Sun, Weiyu Cheng, Wenkai Li, Xiangjun Song, Xiao Su, Xiaodong Han, Xinjie Zhang, Xinzhu Hou, Xu~Min, Xun Zou, Xuyang Shen, Yan Gong, Yingjie Zhu, Yipeng Zhou, Yiran Zhong, Yongyi Hu, Yuanxiang Fan, Yue Yu, Yufeng Yang, Yuhao Li, Yunan Huang, Yunji Li, Yunpeng Huang, Yunzhi Xu, Yuxin Mao, Zehan Li, Zekang Li, Zewei Tao, Zewen Ying, Zhaoyang Cong, Zhen
  Qin, Zhenhua Fan, Zhihang Yu, Zhuo Jiang, and Zijia Wu. 2025.
\newblock \href {https://arxiv.org/abs/2501.08313} {Minimax-01: Scaling foundation models with lightning attention}.
\newblock \emph{Preprint}, arXiv:2501.08313.

\bibitem[{Muennighoff et~al.(2024)Muennighoff, Soldaini, Groeneveld, Lo, Morrison, Min, Shi, Walsh, Tafjord, Lambert et~al.}]{muennighoff2024olmoe}
Niklas Muennighoff, Luca Soldaini, Dirk Groeneveld, Kyle Lo, Jacob Morrison, Sewon Min, Weijia Shi, Pete Walsh, Oyvind Tafjord, Nathan Lambert, et~al. 2024.
\newblock Olmoe: Open mixture-of-experts language models.
\newblock \emph{arXiv preprint arXiv:2409.02060}.

\bibitem[{Peng et~al.(2023)Peng, Alcaide, Anthony, Albalak, Arcadinho, Biderman, Cao, Cheng, Chung, Derczynski, Du, Grella, Gv, He, Hou, Kazienko, Kocon, Kong, Koptyra, Lau, Lin, Mantri, Mom, Saito, Song, Tang, Wind, Wo{\'z}niak, Zhang, Zhou, Zhu, and Zhu}]{peng-etal-2023-rwkv}
Bo~Peng, Eric Alcaide, Quentin Anthony, Alon Albalak, Samuel Arcadinho, Stella Biderman, Huanqi Cao, Xin Cheng, Michael Chung, Leon Derczynski, Xingjian Du, Matteo Grella, Kranthi Gv, Xuzheng He, Haowen Hou, Przemyslaw Kazienko, Jan Kocon, Jiaming Kong, Bart{\l}omiej Koptyra, Hayden Lau, Jiaju Lin, Krishna Sri~Ipsit Mantri, Ferdinand Mom, Atsushi Saito, Guangyu Song, Xiangru Tang, Johan Wind, Stanis{\l}aw Wo{\'z}niak, Zhenyuan Zhang, Qinghua Zhou, Jian Zhu, and Rui-Jie Zhu. 2023.
\newblock \href {https://doi.org/10.18653/v1/2023.findings-emnlp.936} {{RWKV}: Reinventing {RNN}s for the transformer era}.
\newblock In \emph{Findings of the Association for Computational Linguistics: EMNLP 2023}, pages 14048--14077, Singapore. Association for Computational Linguistics.

\bibitem[{Peng et~al.(2024)Peng, Goldstein, Anthony, Albalak, Alcaide, Biderman, Cheah, Ferdinan, Hou, Kazienko et~al.}]{peng2024eagle}
Bo~Peng, Daniel Goldstein, Quentin Anthony, Alon Albalak, Eric Alcaide, Stella Biderman, Eugene Cheah, Teddy Ferdinan, Haowen Hou, Przemys{\l}aw Kazienko, et~al. 2024.
\newblock Eagle and finch: Rwkv with matrix-valued states and dynamic recurrence.
\newblock \emph{arXiv preprint arXiv:2404.05892}.

\bibitem[{Peng et~al.(2025)Peng, Zhang, Goldstein, Alcaide, Hou, Lu, Merrill, Song, Tan, Utpala et~al.}]{peng2025rwkv}
Bo~Peng, Ruichong Zhang, Daniel Goldstein, Eric Alcaide, Haowen Hou, Janna Lu, William Merrill, Guangyu Song, Kaifeng Tan, Saiteja Utpala, et~al. 2025.
\newblock Rwkv-7" goose" with expressive dynamic state evolution.
\newblock \emph{arXiv preprint arXiv:2503.14456}.

\bibitem[{Pouransari et~al.(2024)Pouransari, Li, Chang, Vasu, Koc, Shankar, and Tuzel}]{pouransari2024dataset}
Hadi Pouransari, Chun-Liang Li, Jen-Hao~Rick Chang, Pavan Kumar~Anasosalu Vasu, Cem Koc, Vaishaal Shankar, and Oncel Tuzel. 2024.
\newblock Dataset decomposition: Faster llm training with variable sequence length curriculum.
\newblock \emph{arXiv preprint arXiv:2405.13226}.

\bibitem[{Qin et~al.(2024{\natexlab{a}})Qin, Li, Sun, Sun, Shen, Han, Wei, Lv, Luo, Qiao, and Zhong}]{qin2024transnormerllm}
Zhen Qin, Dong Li, Weigao Sun, Weixuan Sun, Xuyang Shen, Xiaodong Han, Yunshen Wei, Baohong Lv, Xiao Luo, Yu~Qiao, and Yiran Zhong. 2024{\natexlab{a}}.
\newblock \href {https://arxiv.org/abs/2307.14995} {Transnormerllm: A faster and better large language model with improved transnormer}.
\newblock \emph{Preprint}, arXiv:2307.14995.

\bibitem[{Qin et~al.(2024{\natexlab{b}})Qin, Shen, Sun, Li, Birchfield, Hartley, and Zhong}]{qin2024unlocking}
Zhen Qin, Xuyang Shen, Weigao Sun, Dong Li, Stan Birchfield, Richard Hartley, and Yiran Zhong. 2024{\natexlab{b}}.
\newblock Unlocking the secrets of linear complexity sequence model from a unified perspective.
\newblock \emph{arXiv preprint arXiv:2405.17383}.

\bibitem[{Qin et~al.(2024{\natexlab{c}})Qin, Sun, Li, Shen, Sun, and Zhong}]{qin2024lightning}
Zhen Qin, Weigao Sun, Dong Li, Xuyang Shen, Weixuan Sun, and Yiran Zhong. 2024{\natexlab{c}}.
\newblock Lightning attention-2: A free lunch for handling unlimited sequence lengths in large language models.
\newblock \emph{arXiv preprint arXiv:2401.04658}.

\bibitem[{Qin et~al.(2024{\natexlab{d}})Qin, Yang, Sun, Shen, Li, Sun, and Zhong}]{qin2024hgrn2}
Zhen Qin, Songlin Yang, Weixuan Sun, Xuyang Shen, Dong Li, Weigao Sun, and Yiran Zhong. 2024{\natexlab{d}}.
\newblock Hgrn2: Gated linear rnns with state expansion.
\newblock \emph{arXiv preprint arXiv:2404.07904}.

\bibitem[{Qin et~al.(2024{\natexlab{e}})Qin, Yang, and Zhong}]{qin2024hierarchically}
Zhen Qin, Songlin Yang, and Yiran Zhong. 2024{\natexlab{e}}.
\newblock Hierarchically gated recurrent neural network for sequence modeling.
\newblock \emph{Advances in Neural Information Processing Systems}, 36.

\bibitem[{Qu et~al.(2024)Qu, Dong, Hu, Zhu, Sun, and Cheng}]{qu2024llama}
Xiaoye Qu, Daize Dong, Xuyang Hu, Tong Zhu, Weigao Sun, and Yu~Cheng. 2024.
\newblock Llama-moe v2: Exploring sparsity of llama from perspective of mixture-of-experts with post-training.
\newblock \emph{arXiv preprint arXiv:2411.15708}.

\bibitem[{Reid et~al.(2024)Reid, Savinov, Teplyashin, Lepikhin, Lillicrap, Alayrac, Soricut, Lazaridou, Firat, Schrittwieser et~al.}]{reid2024gemini}
Machel Reid, Nikolay Savinov, Denis Teplyashin, Dmitry Lepikhin, Timothy Lillicrap, Jean-baptiste Alayrac, Radu Soricut, Angeliki Lazaridou, Orhan Firat, Julian Schrittwieser, et~al. 2024.
\newblock Gemini 1.5: Unlocking multimodal understanding across millions of tokens of context.
\newblock \emph{arXiv preprint arXiv:2403.05530}.

\bibitem[{Ren et~al.(2024)Ren, Liu, Lu, Shen, Liang, and Chen}]{ren2024samba}
Liliang Ren, Yang Liu, Yadong Lu, Yelong Shen, Chen Liang, and Weizhu Chen. 2024.
\newblock Samba: Simple hybrid state space models for efficient unlimited context language modeling.
\newblock \emph{arXiv preprint arXiv:2406.07522}.

\bibitem[{Roller et~al.(2021)Roller, Sukhbaatar, Weston et~al.}]{roller2021hash}
Stephen Roller, Sainbayar Sukhbaatar, Jason Weston, et~al. 2021.
\newblock Hash layers for large sparse models.
\newblock \emph{Advances in Neural Information Processing Systems}, 34:17555--17566.

\bibitem[{Sakaguchi et~al.(2019)Sakaguchi, Bras, Bhagavatula, and Choi}]{sakaguchi2019winogrande}
Keisuke Sakaguchi, Ronan~Le Bras, Chandra Bhagavatula, and Yejin Choi. 2019.
\newblock Winogrande: An adversarial winograd schema challenge at scale.
\newblock \emph{arXiv preprint arXiv:1907.10641}.

\bibitem[{Schlag et~al.(2021)Schlag, Irie, and Schmidhuber}]{Schlag2021LinearTA}
Imanol Schlag, Kazuki Irie, and J{\"u}rgen Schmidhuber. 2021.
\newblock Linear transformers are secretly fast weight programmers.
\newblock In \emph{International Conference on Machine Learning}.

\bibitem[{Shazeer et~al.(2017)Shazeer, Mirhoseini, Maziarz, Davis, Le, Hinton, and Dean}]{shazeer2017outrageously}
Noam Shazeer, Azalia Mirhoseini, Krzysztof Maziarz, Andy Davis, Quoc Le, Geoffrey Hinton, and Jeff Dean. 2017.
\newblock Outrageously large neural networks: The sparsely-gated mixture-of-experts layer.
\newblock \emph{arXiv preprint arXiv:1701.06538}.

\bibitem[{Shen et~al.(2024)Shen, Guo, Cai, and Qin}]{shen2024jetmoe}
Yikang Shen, Zhen Guo, Tianle Cai, and Zengyi Qin. 2024.
\newblock Jetmoe: Reaching llama2 performance with 0.1 m dollars.
\newblock \emph{arXiv preprint arXiv:2404.07413}.

\bibitem[{Soboleva et~al.(2023)Soboleva, Al-Khateeb, Myers, Steeves, Hestness, and Dey}]{cerebras2023slimpajama}
Daria Soboleva, Faisal Al-Khateeb, Robert Myers, Jacob~R Steeves, Joel Hestness, and Nolan Dey. 2023.
\newblock \href {https://huggingface.co/datasets/cerebras/SlimPajama-627B} {{SlimPajama: A 627B token cleaned and deduplicated version of RedPajama}}.

\bibitem[{Sun et~al.(2025)Sun, Lan, Zhong, Qu, and Cheng}]{sun2025lasp}
Weigao Sun, Disen Lan, Yiran Zhong, Xiaoye Qu, and Yu~Cheng. 2025.
\newblock Lasp-2: Rethinking sequence parallelism for linear attention and its hybrid.
\newblock \emph{arXiv preprint arXiv:2502.07563}.

\bibitem[{Sun et~al.(2024{\natexlab{a}})Sun, Qin, Li, Shen, Qiao, and Zhong}]{sun2024linear}
Weigao Sun, Zhen Qin, Dong Li, Xuyang Shen, Yu~Qiao, and Yiran Zhong. 2024{\natexlab{a}}.
\newblock Linear attention sequence parallelism.
\newblock \emph{arXiv preprint arXiv:2404.02882}.

\bibitem[{Sun et~al.(2024{\natexlab{b}})Sun, Li, Dalal, Xu, Vikram, Zhang, Dubois, Chen, Wang, Koyejo et~al.}]{sun2024learning}
Yu~Sun, Xinhao Li, Karan Dalal, Jiarui Xu, Arjun Vikram, Genghan Zhang, Yann Dubois, Xinlei Chen, Xiaolong Wang, Sanmi Koyejo, et~al. 2024{\natexlab{b}}.
\newblock Learning to (learn at test time): Rnns with expressive hidden states.
\newblock \emph{arXiv preprint arXiv:2407.04620}.

\bibitem[{Sun et~al.(2023)Sun, Dong, Huang, Ma, Xia, Xue, Wang, and Wei}]{sun2023retentive}
Yutao Sun, Li~Dong, Shaohan Huang, Shuming Ma, Yuqing Xia, Jilong Xue, Jianyong Wang, and Furu Wei. 2023.
\newblock Retentive network: A successor to transformer for large language models.
\newblock \emph{arXiv preprint arXiv:2307.08621}.

\bibitem[{Team et~al.(2024)Team, Lenz, Arazi, Bergman, Manevich, Peleg, Aviram, Almagor, Fridman, Padnos et~al.}]{team2024jamba}
Jamba Team, Barak Lenz, Alan Arazi, Amir Bergman, Avshalom Manevich, Barak Peleg, Ben Aviram, Chen Almagor, Clara Fridman, Dan Padnos, et~al. 2024.
\newblock Jamba-1.5: Hybrid transformer-mamba models at scale.
\newblock \emph{arXiv preprint arXiv:2408.12570}.

\bibitem[{Vaswani et~al.(2017)Vaswani, Shazeer, Parmar, Uszkoreit, Jones, Gomez, Kaiser, and Polosukhin}]{vaswani2017attention}
Ashish Vaswani, Noam Shazeer, Niki Parmar, Jakob Uszkoreit, Llion Jones, Aidan~N Gomez, {\L}ukasz Kaiser, and Illia Polosukhin. 2017.
\newblock Attention is all you need.
\newblock \emph{Advances in neural information processing systems}, 30.

\bibitem[{Waleffe et~al.(2024)Waleffe, Byeon, Riach, Norick, Korthikanti, Dao, Gu, Hatamizadeh, Singh, Narayanan et~al.}]{waleffe2024empirical}
Roger Waleffe, Wonmin Byeon, Duncan Riach, Brandon Norick, Vijay Korthikanti, Tri Dao, Albert Gu, Ali Hatamizadeh, Sudhakar Singh, Deepak Narayanan, et~al. 2024.
\newblock An empirical study of mamba-based language models.
\newblock \emph{arXiv preprint arXiv:2406.07887}.

\bibitem[{Wang et~al.(2024)Wang, Gao, Zhao, Sun, and Dai}]{wang2024auxiliary}
Lean Wang, Huazuo Gao, Chenggang Zhao, Xu~Sun, and Damai Dai. 2024.
\newblock \href {https://arxiv.org/abs/2408.15664} {Auxiliary-loss-free load balancing strategy for mixture-of-experts}.
\newblock \emph{Preprint}, arXiv:2408.15664.

\bibitem[{Yang et~al.(2024{\natexlab{a}})Yang, Yang, Hui, Zheng, Yu, Zhou, Li, Li, Liu, Huang, Dong, Wei, Lin, Tang, Wang, Yang, Tu, Zhang, Ma, Xu, Zhou, Bai, He, Lin, Dang, Lu, Chen, Yang, Li, Xue, Ni, Zhang, Wang, Peng, Men, Gao, Lin, Wang, Bai, Tan, Zhu, Li, Liu, Ge, Deng, Zhou, Ren, Zhang, Wei, Ren, Fan, Yao, Zhang, Wan, Chu, Liu, Cui, Zhang, and Fan}]{qwen2}
An~Yang, Baosong Yang, Binyuan Hui, Bo~Zheng, Bowen Yu, Chang Zhou, Chengpeng Li, Chengyuan Li, Dayiheng Liu, Fei Huang, Guanting Dong, Haoran Wei, Huan Lin, Jialong Tang, Jialin Wang, Jian Yang, Jianhong Tu, Jianwei Zhang, Jianxin Ma, Jin Xu, Jingren Zhou, Jinze Bai, Jinzheng He, Junyang Lin, Kai Dang, Keming Lu, Keqin Chen, Kexin Yang, Mei Li, Mingfeng Xue, Na~Ni, Pei Zhang, Peng Wang, Ru~Peng, Rui Men, Ruize Gao, Runji Lin, Shijie Wang, Shuai Bai, Sinan Tan, Tianhang Zhu, Tianhao Li, Tianyu Liu, Wenbin Ge, Xiaodong Deng, Xiaohuan Zhou, Xingzhang Ren, Xinyu Zhang, Xipin Wei, Xuancheng Ren, Yang Fan, Yang Yao, Yichang Zhang, Yu~Wan, Yunfei Chu, Yuqiong Liu, Zeyu Cui, Zhenru Zhang, and Zhihao Fan. 2024{\natexlab{a}}.
\newblock Qwen2 technical report.
\newblock \emph{arXiv preprint arXiv:2407.10671}.

\bibitem[{Yang et~al.(2024{\natexlab{b}})Yang, Kautz, and Hatamizadeh}]{yang2024gated}
Songlin Yang, Jan Kautz, and Ali Hatamizadeh. 2024{\natexlab{b}}.
\newblock Gated delta networks: Improving mamba2 with delta rule.
\newblock \emph{arXiv preprint arXiv:2412.06464}.

\bibitem[{Yang et~al.(2023)Yang, Wang, Shen, Panda, and Kim}]{yang2023gated}
Songlin Yang, Bailin Wang, Yikang Shen, Rameswar Panda, and Yoon Kim. 2023.
\newblock Gated linear attention transformers with hardware-efficient training.
\newblock \emph{arXiv preprint arXiv:2312.06635}.

\bibitem[{Yang et~al.(2024{\natexlab{c}})Yang, Wang, Zhang, Shen, and Kim}]{yang2024parallelizing}
Songlin Yang, Bailin Wang, Yu~Zhang, Yikang Shen, and Yoon Kim. 2024{\natexlab{c}}.
\newblock Parallelizing linear transformers with the delta rule over sequence length.
\newblock \emph{arXiv preprint arXiv:2406.06484}.

\bibitem[{Zellers et~al.(2019)Zellers, Holtzman, Bisk, Farhadi, and Choi}]{zellers2019hellaswag}
Rowan Zellers, Ari Holtzman, Yonatan Bisk, Ali Farhadi, and Yejin Choi. 2019.
\newblock Hellaswag: Can a machine really finish your sentence?
\newblock In \emph{Proceedings of the 57th Annual Meeting of the Association for Computational Linguistics}.

\bibitem[{Zeng et~al.(2022)Zeng, Li, Wu, Liu, Liu, Yu, and Ma}]{zeng2022boosting}
Jinle Zeng, Min Li, Zhihua Wu, Jiaqi Liu, Yuang Liu, Dianhai Yu, and Yanjun Ma. 2022.
\newblock Boosting distributed training performance of the unpadded bert model.
\newblock \emph{arXiv preprint arXiv:2208.08124}.

\bibitem[{Zhai et~al.(2023)Zhai, Jiang, Wang, Jia, Zhang, Chen, Liu, and Zhu}]{zhai2023bytetransformer}
Yujia Zhai, Chengquan Jiang, Leyuan Wang, Xiaoying Jia, Shang Zhang, Zizhong Chen, Xin Liu, and Yibo Zhu. 2023.
\newblock Bytetransformer: A high-performance transformer boosted for variable-length inputs.
\newblock In \emph{2023 IEEE International Parallel and Distributed Processing Symposium (IPDPS)}, pages 344--355. IEEE.

\bibitem[{Zhang et~al.(2024)Zhang, Yang, Zhu, Zhang, Cui, Wang, Wang, Freda~Shi, Zhou, and Fu}]{zhang2024gsa}
Yu~Zhang, Songlin Yang, Ruijie Zhu, Yue Zhang, Leyang Cui, Yiqiao Wang, Bolun Wang, Wei~Bi Freda~Shi, Bailin~Wang, Peng Zhou, and Guohong Fu. 2024.
\newblock Gated slot attention for efficient linear-time sequence modeling.
\newblock \emph{arXiv preprint arXiv:2409.07146}.

\bibitem[{Zhou et~al.(2022)Zhou, Lei, Liu, Du, Huang, Zhao, Dai, Le, Laudon et~al.}]{zhou2022mixture}
Yanqi Zhou, Tao Lei, Hanxiao Liu, Nan Du, Yanping Huang, Vincent Zhao, Andrew~M Dai, Quoc~V Le, James Laudon, et~al. 2022.
\newblock Mixture-of-experts with expert choice routing.
\newblock \emph{Advances in Neural Information Processing Systems}, 35:7103--7114.

\bibitem[{Zhu et~al.(2024{\natexlab{a}})Zhu, Dong, Qu, Ruan, Chen, and Cheng}]{zhu2024dynamic}
Tong Zhu, Daize Dong, Xiaoye Qu, Jiacheng Ruan, Wenliang Chen, and Yu~Cheng. 2024{\natexlab{a}}.
\newblock Dynamic data mixing maximizes instruction tuning for mixture-of-experts.
\newblock \emph{arXiv preprint arXiv:2406.11256}.

\bibitem[{Zhu et~al.(2024{\natexlab{b}})Zhu, Qu, Dong, Ruan, Tong, He, and Cheng}]{zhu2024llama}
Tong Zhu, Xiaoye Qu, Daize Dong, Jiacheng Ruan, Jingqi Tong, Conghui He, and Yu~Cheng. 2024{\natexlab{b}}.
\newblock Llama-moe: Building mixture-of-experts from llama with continual pre-training.
\newblock \emph{arXiv preprint arXiv:2406.16554}.

\end{thebibliography}

\newpage
\appendix
\section{Appendix}
%

\subsection{Related Work}

\subsubsection{Mixture-of-Experts}
MoE~\citep{Jacobs_Jordan_Nowlan_Hinton_1991,cai2024survey,lu2025twin} is gaining increasing attention in the development of large language models (LLMs) due to its ability to scale model size while maintaining computational efficiency. Its key strength lies in the sparse activation of experts and routing mechanisms, enabling a better balance between model performance and training cost. The effectiveness of MoE in modern deep learning was first demonstrated in~\citet{shazeer2017outrageously}, where an MoE layer was introduced between LSTM layers, resulting in state-of-the-art performance on language modeling and machine translation benchmarks. Following this, the MoE layer was incorporated into the Transformer architecture, replacing the feed-forward network (FFN) layers. GShard~\citep{lepikhin2020gshard} applied MoE to Transformers, significantly improving machine translation across 100 languages. Switch Transformers~\citep{fedus2022switch} further scaled model size to trillions of parameters, using a simplified and efficient MoE layer design.
However, training MoE models often leads to load imbalance, where only a few experts are heavily utilized, leaving others underutilized \cite{lewis2021base,wang2024auxiliary,zhu2024dynamic,du2025mom}. To address this, several strategies have been developed to optimize MoE training. These include the BASE layer~\citep{lewis2021base}, the HASH layer~\citep{roller2021hash}, and Expert Choice~\citep{zhou2022mixture}, all of which aim to maximize model capacity utilization. MoE architectures have been widely adopted in industry-leading models, such as Gemini-1.5~\citep{reid2024gemini} and reportedly GPT-4~\citep{gpt4}. Other notable examples of LLMs incorporating MoE techniques include Mixtral~\citep{jiang2024mixtral}, DeepSeek V2~\citep{liu2024deepseek}, Qwen2~\citep{qwen2}, JetMoE~\citep{shen2024jetmoe}, Jamba~\citep{team2024jamba}, and OLMoE~\citep{muennighoff2024olmoe}.
Despite the advances in MoE, most research has focused on improving FFN layers and routers, while attention mechanisms have remained largely unchanged. There is still much room for exploring how to enhance the efficiency of MoE models by evolving their attention layers.

\subsubsection{Linear Sequence Modeling}
\textbf{Linear Attention.}
Linear attention encompasses a set of techniques aimed at calculating attention outputs using the "right-product kernel trick," which first computes key-value products, thereby avoiding the quadratic complexity associated with query-key computations. Vanilla linear attention~\citep{katharopoulos2020transformers} replaces the $\operatorname{Softmax}$ attention~\citep{vaswani2017attention} with kernel methods, reducing the computational complexity to linear in relation to sequence length. Building on this, various extensions of linear attention have emerged. For example, TransNormerLLM~\citep{qin2024transnormerllm} introduces Lightning Attention, an optimized linear attention mechanism that speeds up processing by enhancing IO operations. Lightning Attention-2~\citep{qin2024lightning} further improves this by separately handling inter- and intra-block computations to fully exploit the advantages of linear attention on autoregressive tasks. RetNet~\citep{sun2023retentive} combines a retention mechanism with attention, offering both parallel training and linear-time inference. Gated Linear Attention (GLA)~\citep{yang2023gated} introduces a data-independent gating mechanism and presents a hardware-efficient algorithm for training. DeltaNet~\citep{Schlag2021LinearTA}, along with its parallelized version~\citep{yang2024parallelizing}, applies a delta-rule-like update to improve performance in long-context scenarios. More recently, Gated Slot Attention (GSA)~\citep{zhang2024gsa}, inspired by GLA, introduces a bounded-memory slot control mechanism within the gated linear attention framework, further boosting performance in tasks requiring strong recall abilities. Recent work Gated DeltaNet~\citep{yang2024gated}, TTT~\citep{sun2024learning}, Titans~\citep{behrouz2024titans} propose new update rules of linear attention mechanism, which are able to learn or memorize at test-time, and achieve better performance.

\textbf{State Space Model.} SSM provides a robust framework for capturing the behavior of sequence modeling within dynamic systems, and has demonstrated itself in the field of linear sequence modeling. Models such as S4~\citep{gu2022efficiently} and its subsequent variants~\citep{gu2022parameterization,gupta2022diagonal} have achieved notable success, particularly in long-range synthetic tasks. A recent example is Mamba~\citep{gu2023mamba}, a representative SSM model that introduces a state selection mechanism. Mamba addresses the limitation of static dynamics in previous methods, arguing that they do not account for input-specific context selection within the hidden state, which is critical for tasks like language modeling. Mamba has shown superior performance compared to Transformers across various model sizes and scales. Mamba has been further refined in its successor, Mamba2~\citep{dao2024transformers}, which integrates a linear attention-like mechanism that improves hardware efficiency during training. Similar to how linear attention uses outer products to expand the state, Mamba2 leverages a state-space duality that enables parallel attention-style computation while maintaining recurrent inference capabilities.

\textbf{Linear RNN.} Traditional RNNs struggle with long-context sequence modeling, largely due to their sequential nature during training, which limits their ability to benefit from scaling laws~\citep{sun2023retentive}. To mitigate these issues, Linear RNNs introduce parallel training capabilities, achieving competitive performance with Transformers of comparable size. RWKV~\citep{peng-etal-2023-rwkv,peng2024eagle} is an example of a large language model based on linear RNNs, designed to effectively manage long-term dependencies. Furthermore, HGRN~\citep{qin2024hierarchically} emphasizes the importance of data-dependent decay mechanisms in enhancing linear RNN performance, showing how tuning decay parameters can improve learning in long-context scenarios. The upgraded HGRN2~\citep{qin2024hgrn2} builds on this by introducing a state expansion mechanism that leverages outer product operations, allowing for better scalability and improved sequence modeling over extended inputs. Both RWKV and HGRN models aim to address the limitations of traditional RNNs for efficient long-sequence modeling.

\subsection{Tensor Parallelism on Linear-MoE}
\label{subsubsec: TP}

The core computation mechanism of LSM modules can be abstracted in the following general form:
\begin{equation}
\begin{aligned}
&\mathbf O=\phi(\mathbf{Q}) (\phi(\mathbf{K})^{\top}\mathbf{V}), \\
&\mathbf{Q}=\mathbf X \mathbf W_Q, \mathbf{K}=\mathbf X \mathbf W_K, \mathbf{V}=\mathbf X \mathbf W_V,
\label{eq: mp_lsm1}
\end{aligned}
\end{equation}
where TP is applied by splitting the matrix multiplications as follows:
\begin{equation}
\begin{aligned}
\mathbf Q&=[\phi(\mathbf X \mathbf W_Q^1), \phi(\mathbf X \mathbf W_Q^2)],\\
\mathbf K&=[\phi(\mathbf X \mathbf W_K^1), \phi(\mathbf X \mathbf W_K^2)],\\
\mathbf V&=\mathbf X [\mathbf W_V^1, \mathbf W_V^2],\\
\mathbf O&=[\mathbf{O_1}, \mathbf{O_2}],
\label{eq: mp_lsm2}
\end{aligned}
\end{equation}
\normalsize
where the weight matrices $\mathbf W_q$, $\mathbf W_k$, and $\mathbf W_v$ are divided along their columns, producing an output matrix $\mathbf O$ that is also split along columns.

The split output $[\mathbf O_1, \mathbf O_2]$ is then multiplied by an output linear weight matrix that is split along its rows, resulting in:
\begin{equation}
\begin{aligned}
\mathbf {O}&=[\mathbf O_1, \mathbf O_2] [\mathbf W_O^1, \mathbf W_O^2]^\top\\
&=\mathbf O_1 \mathbf W_O^1 + \mathbf O_2 \mathbf W_O^2,
\label{eq: mp_output}
\end{aligned}
\end{equation}
which produces a unified output.

As with TP in standard attention, TP for LSM modules introduces an \textit{all-reduce} collective communication operation during both the forward and backward passes. In practical terms, this all-reduce operation is implemented via two separate steps: all-gather and reduce-scatter, which together functionally achieve the same result as a single all-reduce.

\subsection{Sequence Parallelism on Linear-MoE}
\label{app: sp}

\begin{algorithm}[H]
\small
    \caption{SP on Linear-MoE w/o Masking}
    \label{algo:LASP2 fw without mask}
    \begin{algorithmic}[1]
    \State{\textbf{Input:} input sequence $\mathbf X$, distributed world size $W$, sequence parallel size $T=W$.}
    \State{Distribute $\mathbf X = [\mathbf X_t]_{1}^{T}$.}
    \For{chunk $t \in \{1, \cdots, T\}$ on ranks $\{1, \cdots, W\}$ \textbf{in parallel}}
        \State{Calculate $\mathbf Q_t=\mathbf{X}_t \mathbf{W}_Q$, $\mathbf  K_t=\mathbf{X}_t \mathbf{W}_K$, $\mathbf V_t =\mathbf{X}_t \mathbf{W}_V$.}
        \State{Compute $\mathbf{M}_{t} = \mathbf K_t^{\top}  \mathbf V_t$.}
        \State{Communicate $[\mathbf{M}_t]_{1}^\top = \texttt{AllGather}([\mathbf{M}_t]_{1}^\top).$}
        \State{Compute $\mathbf{M}_{1:{T}} = \texttt{Sum}([\mathbf{M}_t]_{1}^{T})$.}
        \State{Compute $\mathbf{O}_{t} =\mathbf Q_t \mathbf {M}_{1:{T}} $.}
      \EndFor
      \State{return $\mathbf O = [\mathbf O_t]_{1}^{T}$.}
\end{algorithmic}
\end{algorithm}

\begin{algorithm}[H]
\small
    \caption{SP on Linear-MoE w/ Masking}
    \label{algo:LASP2 fw with mask}
    \begin{algorithmic}[1]
    \State{\textbf{Input:} input sequence $\mathbf X$, distributed world size $W$, sequence parallel size $T=W$.}
    \State{Distribute $\mathbf X = [\mathbf X_t]_{1}^{T}$.}
     \State{Initialize mask matrix $\mathbf \Psi$, where $\mathbf \Psi_{ij} = 1$ if $i \geq j$ and $\mathbf \Psi_{ij} = -\infty$ if $i < j$.}
    \For{chunk $t \in \{1, \cdots, T\}$ on ranks $\{1, \cdots, W\}$ \textbf{in parallel}}
        \State{Calculate $\mathbf Q_t=\mathbf{X}_t \mathbf{W}_Q$, $\mathbf  K_t=\mathbf{X}_t \mathbf{W}_K$, $\mathbf V_t =\mathbf{X}_t \mathbf{W}_V$.}
        \State{Compute $\mathbf{M}_{t} = (\mathbf K_t)^{\top}  \mathbf V_t$.}
        \State{{Communicate $[\mathbf{M}_t]_{1}^\top = \texttt{AllGather}([\mathbf{M}_t]_{1}^\top)$.}}
        \State{{Compute $\mathbf O_{\mathrm{t, intra}}= [(\mathbf Q_t \mathbf K_t^{\top }) \odot \mathbf \Psi]\mathbf V_t$.}}
        \State{Compute prefix sum $$\mathbf{M}_{1:{t-1}} = \texttt{PrefixSum}([\mathbf{M}_t]_{1}^{t-1}).$$}
        \State{Compute $\mathbf{O}_{\mathrm{t, inter}} = \mathbf Q_t \mathbf{M}_{1:{t-1}}$.}
        \State{Compute $\mathbf O_t=\mathbf O_{\mathrm{t, intra}}+ \mathbf{O}_{\mathrm{t, inter}}$.}
      \EndFor
      \State{return $\mathbf O = [\mathbf O_t]_{1}^{T}$.}
\end{algorithmic}
\end{algorithm}

\subsection{Additional Experiments}
\label{app: additional exp}

\begin{table*}[t]
\centering
\small
\resizebox{\linewidth}{!}{
\begin{tabular}{l|ll|cccccc|cc}
\toprule
\multirow{2}{*}{\textbf{Scale}} & \multirow{2}{*}{\textbf{Model}} & \multirow{2}{*}{\textbf{\shortstack{LSM \\ Instance}}} & \textbf{PIQA} &  \textbf{Hella.} & \textbf{Wino.} & \textbf{ARC-e} &  \textbf{ARC-c} & \textbf{MMLU} & \textbf{Avg.} & \textbf{Avg.} \\
\cmidrule(lr){4-11} 
& & & acc $\uparrow$ &  acc\_norm $\uparrow$  & acc $\uparrow$  & acc $\uparrow$ & acc\_norm $\uparrow$ & acc(5-shot) $\uparrow$ & $\uparrow$ & (no MMLU)$\uparrow$  \\
\midrule
& Baseline & Attention & 55.77 & 27.10 & 50.83 & 33.04 & 23.21 & 23.24 & 35.53 &37.99  \\ 
\cmidrule(lr){2-11}
\multirow{11}{*}{\rotatebox{0}{\shortstack{A0.3B-2B \\ 15B Tokens}}}
& \multirow{6}{*}{\shortstack{Pure}} 
& BLA & 64.42 & 33.41 & 49.01 & 48.15 & 24.32 & 26.32 & 40.94 & 43.86 \\
& & Retention &62.08 & 29.14 & 50.75 & 42.72 & 21.50 & 23.12 & 39.60 &43.39 \\
& & GLA & 65.56 & 35.29 & 50.67 & 47.81 & 23.04 & 24.85 & 41.20 &44.47 \\
& & Mamba2 & 66.97 & 37.79 & 50.20 & 49.12 & 24.74& 25.85 & 42.45 &45.76 \\
& & HGRN2 & 52.50 & 26.37 & 49.01 & 24.83 & 27.65 & 25.10 & 34.24 & 36.07 \\
\cmidrule(lr){2-11}
& \multirow{6}{*}{\shortstack{Hybrid}} 
& BLA &66.76 & 37.16 & 49.96 & 49.62 & 24.74& 25.64 & 42.31 & 45.65 \\
& & Retention & 66.21 & 36.06 & 51.54 & 47.18 & 24.91& 23.71 & 41.60 & 45.18 \\ 
& & GLA &67.71 & 38.62 & 49.72 & 50.51 & 26.02& 25.05 & 42.94 & 46.52 \\  
& & Mamba2 & 66.38 & 38.81 & 51.30 & 50.17 & 24.91 & 24.61 & 42.70 & 46.31 \\
& & HGRN2 &  66.27 & 36.79 & 51.46 & 48.82 & 25.43 & 23.19 & 41.99 & 45.75 \\
\bottomrule
\end{tabular}}
\addtolength{\tabcolsep}{2.5pt}    
\centering
\caption{\textbf{A0.3B-2B Evaluation Results on Language Modeling Benchmarks (No Data Corruption).} All models are pretrained from scratch on the same 15B subset of the SlimPajama dataset with the Qwen2 tokenizer. No benchmark data corruption in the pretraining dataset. The A0.3B-2B hybrid models have a stack as "LLLNLLLNLLLN", where "L" represents the Linear-MoE layer, and "N" represents the normal MoE transformer layer.} 
\label{tab:eval_results_A03}
\end{table*}

\begin{table*}[t]
\centering
\small
\resizebox{\linewidth}{!}{
\begin{tabular}{l|ll|cccccc|cc}
\toprule
\multirow{2}{*}{\textbf{Scale}} & \multirow{2}{*}{\textbf{Model}} & \multirow{2}{*}{\textbf{\shortstack{LSM \\ Instance}}} & \textbf{PIQA} &  \textbf{Hella.} & \textbf{Wino.} & \textbf{ARC-e} &  \textbf{ARC-c} & \textbf{MMLU} & \textbf{Avg.} & \textbf{Avg.} \\
\cmidrule(lr){4-11} 
& & & acc $\uparrow$ &  acc\_norm $\uparrow$  & acc $\uparrow$  & acc $\uparrow$ & acc\_norm $\uparrow$ & acc(5-shot) $\uparrow$ & $\uparrow$ & (no MMLU)$\uparrow$  \\
\midrule
\multirow{3}{*}{\rotatebox{0}{\shortstack{A1B-7B \\ 100B Tokens}}}
& \multirow{3}{*}{\shortstack{Pure}} 
& BLA  & 66.65 & 37.74 & 50.12 & 50.80 & 24.23 & 23.71 & 42.21 & 45.91 \\
& & GLA & 68.17 & 43.51 & 51.22 & 52.48 & 25.09 & 24.83 & 44.22 & 48.09  \\
& & Mamba2 & 69.21 & 41.86 & 51.46 & 52.86 & 25.17 & 23.66 & 44.04 & 48.11 \\
\bottomrule
\end{tabular}}
\addtolength{\tabcolsep}{2.5pt}    
\centering
\caption{\textbf{A1B-7B Evaluation Results on Language Modeling Benchmarks (No Data Corruption).} All models are pretrained from scratch on the same 15B subset of the SlimPajama dataset with the Qwen2 tokenizer. No benchmark data corruption in the pretraining dataset.} 
\label{tab:eval_results_A1}
\end{table*}

\begin{figure}[H]
    \centering
    \includegraphics[width=0.9\columnwidth]{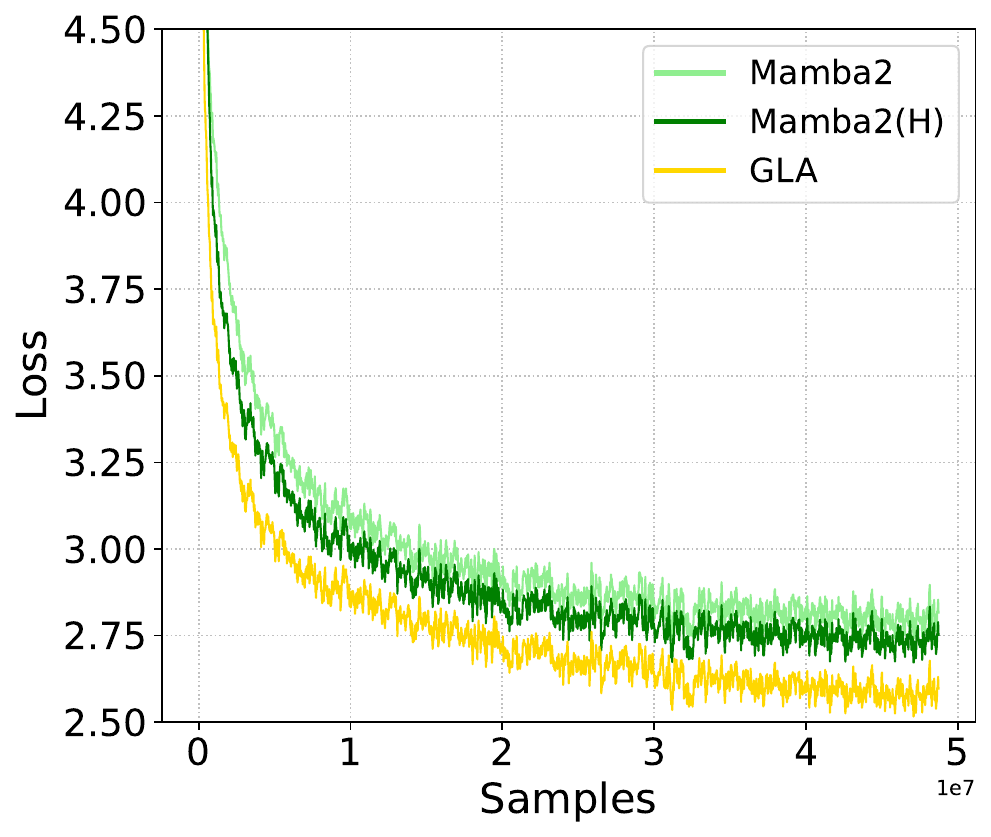}
    \caption{\textbf{Training Loss Curves of A1B-7B Model Instances.}}
    \label{fig: loss-A1B}
\end{figure}


\subsection{Datasets and Benchmarks}
We pretrain all the models on a portion of the SlimPajama dataset which is sampled to approximately 100 billion tokens.

\begin{itemize}
    \item \textbf{SlimPajama} \cite{cerebras2023slimpajama} is a high-quality, optimized subset of the RedPajama dataset, designed for large-scale language model training. It includes diverse text sources such as Common Crawl, Wikipedia, books, and GitHub code, with a primary focus on English. The dataset is cleaned, deduplicated, and optimized for efficiency and performance.
\end{itemize}
For the benchmark, we tested on these tasks:

\begin{itemize}
    \item \textbf{PiQA} \cite{Bisk2020piqa}: A dataset focused on physical commonsense reasoning in English with 3084 test samples. The text consists of everyday tasks and scenarios, requiring models to determine the most practical way to perform an action. The data is sourced from crowdsourced descriptions, reflecting a broad range of common human experiences.
    \item \textbf{ARC-Easy \& ARC-Challenge} \cite{clark2018think}: A set of multiple-choice science questions in English, sourced from standardized exams and educational materials with 2376 and 1172 test samples. The dataset represents the domain of elementary and high school science, with questions authored by educators and test designers. ARC-Easy includes straightforward questions, while ARC-Challenge contains more difficult ones that require advanced reasoning.
    \item \textbf{HellaSwag} \cite{zellers2019hellaswag}: An English-language dataset designed for commonsense reasoning, where models must choose the most plausible continuation of a sentence. The text is derived from activity descriptions (e.g., WikiHow), covering everyday scenarios. The dataset was constructed adversarially to be challenging for language models. It has 10003 test samples.
    \item \textbf{WinoGrande} \cite{sakaguchi2019winogrande}: A large-scale English dataset for commonsense reasoning, based on the Winograd Schema Challenge with 1267 test samples. It tests pronoun resolution in ambiguous contexts, with sentences sourced and refined through crowdsourcing. The dataset aims to reduce annotation biases by diversifying sentence structures and topics.
    \item \textbf{MMLU} \cite{li2023cmmlu}: The MMLU (Massive Multitask Language Understanding) dataset is a comprehensive benchmark designed to evaluate AI models' general knowledge across a wide range of subjects and languages. It comprises 57 distinct categories, spanning elementary-level knowledge to advanced professional topics such as law, physics, history, and computer science. The dataset has been translated into 14 languages using professional human translators, ensuring high-quality and accurate translations. This multilingual approach aims to improve the inclusivity and effectiveness of AI models across different linguistic communities.
\end{itemize}

All datasets used in this work are publicly available and have been released by their original creators, who are responsible for ensuring privacy protection. These datasets are used in accordance with their respective licenses and intended purposes. No modifications or derivative datasets have been created.


\end{document}